\documentclass[11pt]{article}
\usepackage{amsmath, amssymb, amsthm}
\usepackage{graphicx}
\usepackage{bm}
\usepackage{colortbl}
\usepackage[rgb, table, dvipsnames]{xcolor}
\usepackage{booktabs}
\usepackage{makecell}
\usepackage{float}
\usepackage{pifont}
\usepackage{todonotes}
\usepackage[toc,page]{appendix}
\usepackage{multirow}
\usepackage{subcaption} 	
\usepackage{amsmath,amsfonts,amsthm,bm,mathtools}
\usepackage{multicol}
\usepackage{lipsum}
\usepackage{geometry}
\usepackage[backend=bibtex, style=numeric-comp, sorting=none, maxbibnames=99]{biblatex}
\usepackage{algorithm}
\usepackage[noend]{algpseudocode}
\usepackage{arydshln}
\usepackage{quotes}
\definecolor{refcolor}{rgb}{0.65,0,0.35}
\usepackage{algorithm}
\usepackage{algpseudocode}
\usepackage{amsmath}
\usepackage[colorlinks=true]{hyperref}
\hypersetup{	
	linkcolor = refcolor,
	citecolor = refcolor,
	filecolor = refcolor,      
	urlcolor = blue,
	citecolor = refcolor
}

\newtheorem{theorem}{Theorem}[section]  %

\graphicspath{ {}{../} }

\title{\bf Detecting Stochasticity in Discrete Signals via Nonparametric Excursion Theorem}

\author{Sunia Tanweer$^{1,2}$\footnote{tanweer1@msu.edu}
\and Firas A.~Khasawneh$^{2}$}

\date{1. Dept. of Mechanical Engineering, Michigan State University \\ 2. Dept. of Computational Mathematics, Science and Engineering, Michigan State University}

\begin{document}
\maketitle

\vspace{-10mm}
\par\noindent\rule{\textwidth}{0.4pt}
\section*{Abstract}
\label{sec:abstract} 
We develop a practical framework for distinguishing diffusive stochastic processes from deterministic signals using only a single discrete time series. Our approach is based on classical excursion and crossing theorems for continuous semimartingales, which correlates number $N_\varepsilon$ of excursions of magnitude at least $\varepsilon$ with the quadratic variation $[X]_T$ of the process. The scaling law holds universally for all continuous semimartingales with finite quadratic variation, including general Ito diffusions with nonlinear or state-dependent volatility, but fails sharply for deterministic systems---thereby providing a theoretically-certfied method of distinguishing between these dynamics, as opposed to the subjective entropy or recurrence based state of the art methods. We construct a robust data-driven diffusion test. The method compares the empirical excursion counts against the theoretical expectation.
The resulting ratio $K(\varepsilon)=N_{\varepsilon}^{\mathrm{emp}}/N_{\varepsilon}^{\mathrm{theory}}$ is then summarized by a log--log slope deviation measuring the $\varepsilon^{-2}$ law that provides a classification into diffusion-like or not. We demonstrate the method on canonical stochastic systems, some periodic and chaotic maps and systems with additive white noise, as well as the stochastic Duffing system. The approach is nonparametric, model-free, and relies only on the universal small-scale structure of continuous semimartingales.

\paragraph{Keywords: }stochastic diffusion, discrete timeseries, quadratic variation, semimartingales, excursion statistics

\section{Introduction}

The problem of determining whether a discrete-time signal arises from deterministic chaos or stochastic dynamics has been studied for decades~\cite{Subramaniyam2015, Mateos2017, Zanin2021}. Earlier approaches, such as correlation dimension, Lyapunov exponents and classical surrogate tests, provided theoretical foundations but are now recognized as fragile under short data, noise and nonstationarity~\cite{Prado2022, Zanin2021}. Modern work therefore emphasizes finite-sample, noise-robust diagnostics operating directly on observed sequences~\cite{Boaretto2021, Zanin2022}.

Recent contributions can be grouped into several methodological families. \emph{Ordinal-pattern and permutation entropy methods} quantify the statistics of rank orderings and typically separate chaotic and stochastic dynamics in
entropy-complexity representations. This includes multiscale extensions, fuzzy permutation entropy, and ANN-assisted permutation-entropy mapping~\cite{Boaretto2021,Zanin2021, Zunino2012}. \emph{Recurrence-based methods} exploit geometric or temporal asymmetries in
recurrence structures, including recent work using recurrence plots for testing stochasticity~\cite{Hirata2021}. \emph{Network-based approaches} map time series to graphs-such as ordinal transition networks and visibility/horizontal visibility graphs-and detect
deterministic signatures via nonuniform transition motifs or deviations from universal stochastic graph statistics~\cite{Olivares2020, Donner2010, Lacasa2008}. \emph{Information-theoretic and high-dimensional extensions} include Local Structure Entropy (LSE), which scales ordinal analysis to correlated or multivariate systems and has shown strong performance in distinguishing chaos from structured stochastic processes~\cite{Zhang2022}. Finally, \emph{machine-learning-based classifiers}, including ANN models built on ordinal features and deep neural networks trained directly on raw signals, have demonstrated excellent discrimination accuracy for short series, though often at the cost of interpretability and dependence on representative
training corpora \cite{Boaretto2021, Zanin2022}.

Despite their empirical success, the existing approaches for distinguishing chaos from stochasticity share a common characteristic in their methodology: they are fundamentally data-driven and diagnostic rather than theory-determined. Ordinal-pattern~\cite{Zanin2021}, permutation-entropy~\cite{Boaretto2021}, recurrence~\cite{Subramaniyam2015}, and network-based~\cite{Olivares2020, Donner2010} methods all rely on auxiliary constructions---such as time-delay embeddings, symbolic partitions, recurrence thresholds, or graph representations---that introduce multiple hyperparameters whose selection is subjective and often system dependent. In practice, classification outcomes may vary substantially with embedding dimension, delay time, scale, or threshold choice, particularly for short, noisy, or nonstationary signals. Symbolic and graphical methods further require subjective interpretation of plots or clustering tendencies, while machine-learning-based classifiers depend on representative training data, architectural choices, and the assumption that test signals belong to the same distributional class as the training corpus. As a result, even highly accurate empirical methods typically lack universality, interpretability, and guarantees of transferability across systems.

These limitations motivate the development of complementary approaches grounded directly in stochastic process theory. In contrast to empirical pattern-based diagnostics, the method proposed here is informed by rigorous results on continuous semimartingales and their excursion statistics~\cite{Perez2023}. It operates directly on the observed time series without embedding, symbolic discretization, graphical representations, or model training, and its theoretical predictions are determined by intrinsic pathwise quantities such as quadratic variation. This framework therefore provides a principled, system-agnostic criterion for detecting stochastic diffusion, offering a theoretically interpretable alternative to existing empirical methodologies. The theory of excursions in random processes is mathematically rich~\cite{excursions2013, Blumenthal1992}, with classical results describing the expected number of level crossings or excursions under various structural assumptions. Foundational contributions include Rice's formula for the expected number of level crossings in stationary Gaussian processes~\cite{Rice1944}, its extensions by Ito~\cite{Ito1963} and Ylvisaker~\cite{Ylvisaker1965}, and the generalizations by Leadbetter and Cryer for both stationary and non-stationary settings~\cite{Leadbetter1965, LeadbetterCryer1965}. 
Further developments address other stochastic processes, such as the work of  Ivanov~\cite{Ivanov1960}, and discrete-time first-order Markov sequences, for which Zekai Sen derived a recursive characterization of the distribution of up-crossings~\cite{Sen1991}.
These classical excursion results establish the statistical backbone underlying level-crossing behavior in stochastic systems. In this work, we will be using the excursion results recently published for Ito processes~\cite{Perez2023}.

\section{Mathematical Background}

\subsection{Stochastic Dynamical systems}
\label{ssec:DynamicalSystems}

Regular differential equations (ODEs) are the workhorse for modeling continuous change. 
They express the rate of change of a system's state at any given time, allowing us to predict its deterministic evolution. 
However, many real-world systems are influenced by random fluctuations, making their behavior inherently stochastic, necessitating the use of stochastic differential equations.

Stochastic differential equations (SDEs) bridge the gap between ODEs and randomness by incorporating terms that introduce randomness into the model. Unlike ODEs with a single solution trajectory, SDEs have a family of possible solutions. Each solution represents a path the system might take with a certain probability. We denote the state of the system at time $t$ by a vector $X_t \in \mathbb{R}^n$. A drift term $\mu_t$ captures the deterministic component of the system's evolution, similar to the right-hand side of an ODE, while a diffusion term $\Omega_t$ multiplying a scalar Brownian motion $W_t$ (a standard Wiener process $W_t \sim N(0, t)$) injects stochasticity into an equation. 

The introduction of the randomness term into the equation leads to two main mathematical interpretations of the resulting SDE---Ito and Stratonovich---which are given according to
\vspace{-24pt}
\begin{subequations} 
\begin{multicols}{2}      
\noindent
    \label{eq:SDE}
  \begin{equation*}
   dX_t = \mu_t\, dt + \Omega_t\, dW_t \quad (\text{Ito});
  \end{equation*} \qquad \qquad
  \begin{equation*}
    dX_t = \mu_t\,dt + \Omega_t \circ dW_t \quad (\text{Stratonovich}).
  \end{equation*}
  \end{multicols}
  \end{subequations}
\vspace{-10pt}
\noindent These two are merely different approaches with different mathematical properties to solve a mathematical equation and generally give different results. 
For example, the Stratonovich interpretation leads to ordinary chain rule formulas under change of variables, while the Ito formula leads to second order terms. 
On the other hand, in contrast to Ito integrals, Stratonovich integrals are not martingales (see Section~\ref{ssec:local_and_semi}), so the Ito interpretation enjoys important computational advantages. 
This work adopts the Ito interpretation only. 

\subsection{Local Martingales and Semimartingales}
\label{ssec:local_and_semi}

Let $(\Omega, \mathcal{F}, \mathbb{P})$ be a filtered probability space and let $\mathcal{F}^* := (\mathcal{F}_t)_{(t \geq 0)}$ be the filtration of $\mathcal{F}$. An $\mathcal{F}^*$-adapted process $M:[0, \infty] \times \Omega \xrightarrow{} \mathbb{R}$ is an $\mathcal{F}^*$-local martingale,  if there exists a sequence of stopping times $\tau_k$ such that
\begin{enumerate}
\item $\tau_k$ are a.s. increasing, i.e. $\mathbb{P} (\tau_k < \tau_{k+1}) = 1$;
     \item The $\tau_k$ are a.s. divergent, i.e. 
        $\mathbb{P}(\lim_{k\xrightarrow{}\infty} {\tau_k} = \infty) = 1$;
    \item The process $M_{t \wedge \tau_k}$ is an 
    $\mathcal{F}^*$-martingale, i.e. for all 
    $s < t$, 
    $\mathbb{E}[M_{t \wedge \tau_k} \mid \mathcal{F}_s] = M_{s \wedge \tau_k}$.
 \end{enumerate}

On the other hand, a semimartingale $X_t$ is a process which can be decomposed into the sum of a local martingale $M_t$ and a finite variation process $A_t$. Semimartingales are the largest class of processes with respect to which the Ito and Stratonovich integrals can be defined. All Ito processes are semimartingales but a Stratonovich may or may not be one.

\subsection{Excursion Theory for Continuous Semimartingales}
Let $(X_t)_{t\in[0,T]}$ be a real-valued continuous semimartingale with 
canonical decomposition
\[
X_t = X_0 + M_t + A_t,
\]
where $X_0$ is the initial value, $M_t$ is a continuous local martingale and $A_t$ is a finite-variation drift. Its quadratic variation is $[X]_T = \langle M\rangle_T$

Let $N_\varepsilon(X)$ denote the number of excursions or oscillations
of amplitude at least $\varepsilon$. Then the following result holds.

\begin{theorem}[Excursion Law for Continuous Semimartingales~\cite{Perez2023}]
\label{thm:excursion}
Let $X$ be a continuous semimartingale with finite quadratic variation 
$[X]_T<\infty$.  
Then
\[
\lim_{\varepsilon\to0}\,
\varepsilon^2 N_\varepsilon(X)
= \frac{[X]_T}{2}
\qquad \text{in }L^1.
\]
Equivalently,
\[
N_\varepsilon(X)
= \frac{[X]_T}{2\varepsilon^2}\bigl(1+o(1)\bigr).
\]
\end{theorem}

This theorem holds for all continuous Ito diffusions
\[
dX_t = \mu(X_t, t)\,dt + \sigma\Omega(X_t, t)\,dW_t,
\]
including nonlinear, multiplicative, and state-dependent noise.
Deterministic systems do \emph{not} satisfy the $\varepsilon^{-2}$ law,
as their quadratic variation is zero in the continuous limit.

\section{Methods}
\subsection{Quadratic Variation Estimation}

Given discrete observations $X_{t_i}$, we estimate the quadratic variation and process-dependent theoretical curve as
\[
\widehat{[X]}_T
=\sum_{i=0}^{n-1}(X_{t_{i+1}}-X_{t_i})^2 \quad \text{and} \quad N^{\mathrm{theory}}_\varepsilon
=
\frac{\widehat{[X]}_T}{2\varepsilon^2}.
\]

Likewise, given the signal, we can also compute the empirical value of the excursion count as $N^{\mathrm{emp}}_\varepsilon$.

\subsection{Excursion Invariant Statistics}

\begin{figure}[!htbp]
\centering
\includegraphics[width=0.65\linewidth]{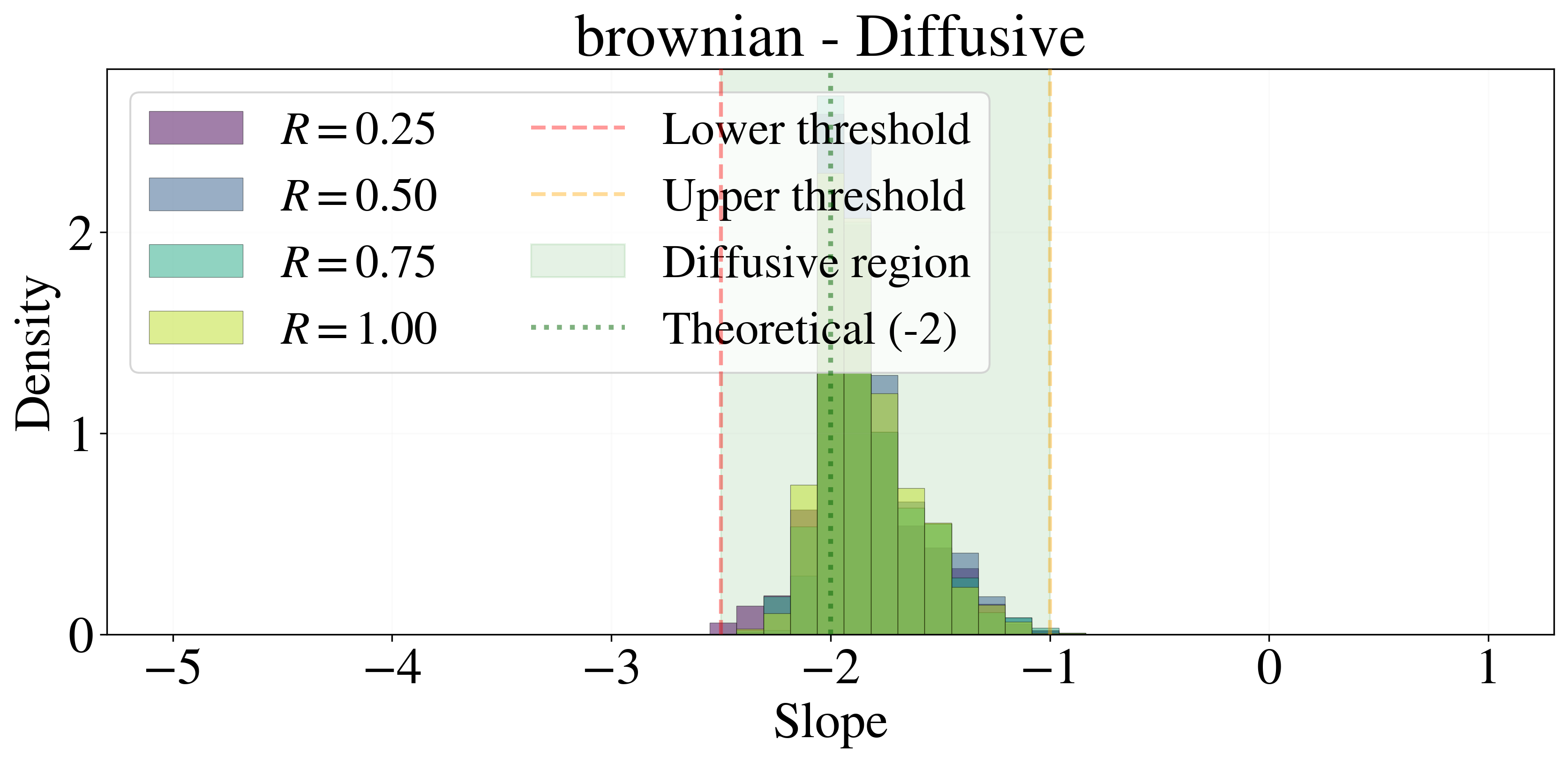}
\caption{Histograms of the fitted log-log slope of the excursion count $N_\varepsilon$ versus $\varepsilon$ for Brownian process.}
\label{fig:hist_brown}
\end{figure}

For a variety of $\varepsilon$, we compute $K(\varepsilon) = {N_\varepsilon^{\mathrm{emp}}}/{N_\varepsilon^{\mathrm{theory}}}.$ A diffusion must satisfy $K(\varepsilon)\approx 1$ across scales. We choose a range of $\varepsilon$ where this is satisfied, and then compute $s$---the slope obtained by fitting $\log N_\varepsilon \;\sim\; s \log\varepsilon.$
An Ito driven stochastic signal must satisfy $s\approx -2$, while a deterministic signal will (typically) take smaller values due to having fewer excursions than a diffusion-driven signal. Fig.~\ref{fig:hist_brown} shows the distribution of the slope for monte carlo simulations for pure Brownian motion. The histogram confirms that the slope is largely concentrated around a value of $-2$, demonstrating that the proposed method can correctly identify Ito diffusions. From the histograms, we also identify upper and lower thresholds, as $-2.5$ and $-1.0$, for the slope to cope with the errors arising due to the discretized and digital nature of real-world signals.

\begin{algorithm}
\caption{Algorithm for Estimation and Classification}
\begin{algorithmic}[1]
\State Compute the quadratic variation estimate $\widehat{[X]}_T$
\State Choose a logarithmic grid of $\varepsilon$
\For{each $\varepsilon$}
    \State Compute the empirical $N_\varepsilon^{\mathrm{emp}}$ via the excursion algorithm
\EndFor
\State Compute the theoretical curve $N_\varepsilon^{\mathrm{theory}}$ using quadratic variation
\State Form the excursion ratio $K(\varepsilon)$ and choose a range of $\varepsilon$ with $K(\varepsilon) \sim 1$
\State Compute $s$ in that range of $\varepsilon$
\State Classify into diffusion if $s \in [-2.5, -1]$; classify non-diffusive otherwise
\end{algorithmic}
\end{algorithm}

\section{Results \& Discussion}

We evaluate the proposed excursion-based diffusion test across a broad class of deterministic and stochastic dynamical systems, with some real-world signals. For each synthetic system, ensembles of discrete time series were generated over a range of observation lengths $T$, sampling intervals $\Delta t$, and signal-to-noise ratios (\textit{$SNR = 10\log({P_{sig}}/P_{noise})$} and \textit{$R
= {\langle \int_0^T \mu(X_t,t)\,dt \rangle}/{\langle \int_0^T \sigma\,\Omega(X_t,t)\,dt \rangle}$} for deterministic and stochastic systems, respectively). Classification accuracy is defined as the fraction of realizations correctly identified as diffusion-like or non-diffusive relative to the known ground truth. Definitions of all systems and parameters used are given in Appendix~\ref{appendix:systems}.

\subsection{Validation on Canonical Diffusion Processes}

\begin{figure}[!htbp]
\centering
\includegraphics[width=0.49\linewidth]{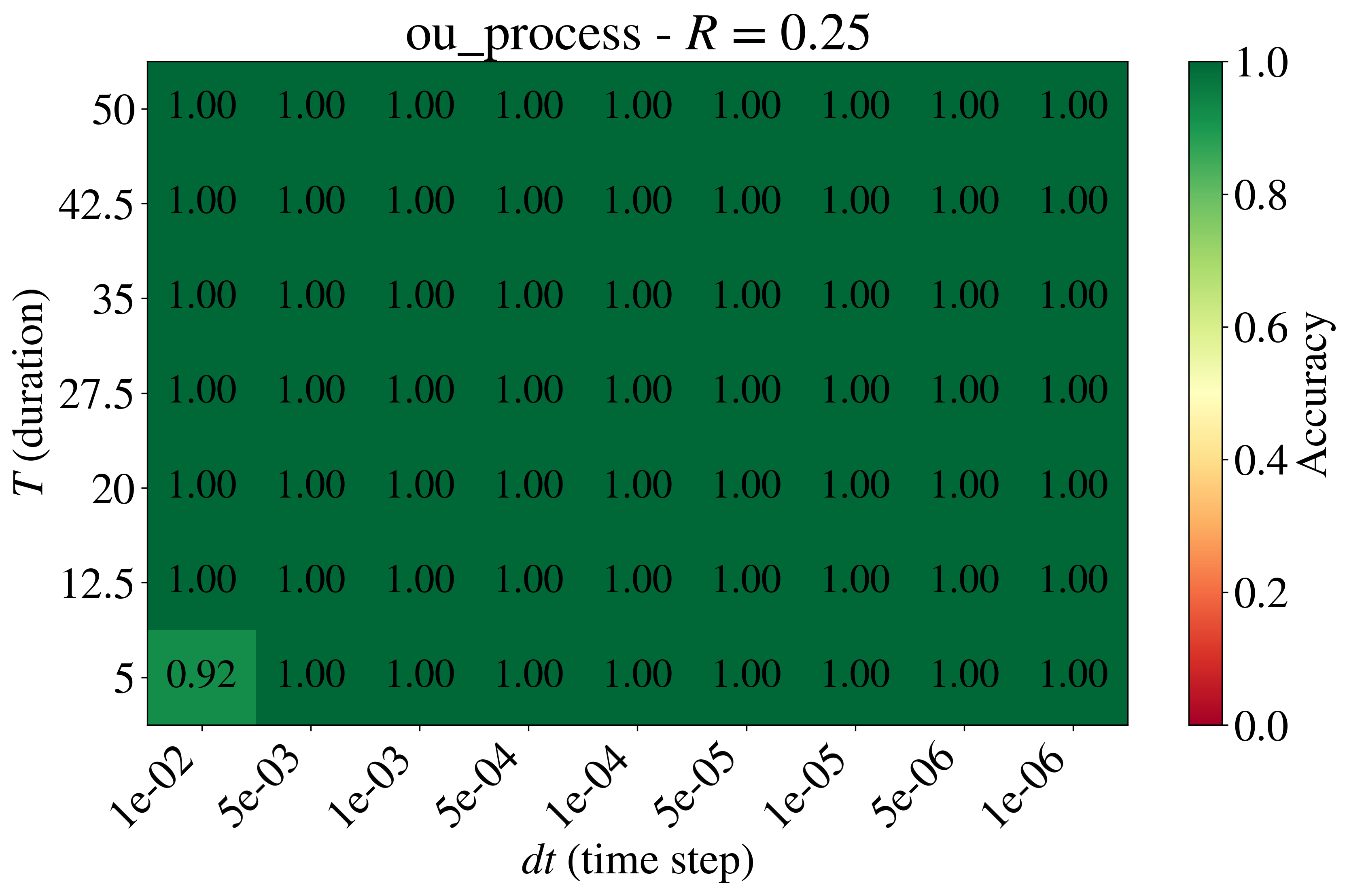}
\includegraphics[width=0.49\linewidth]{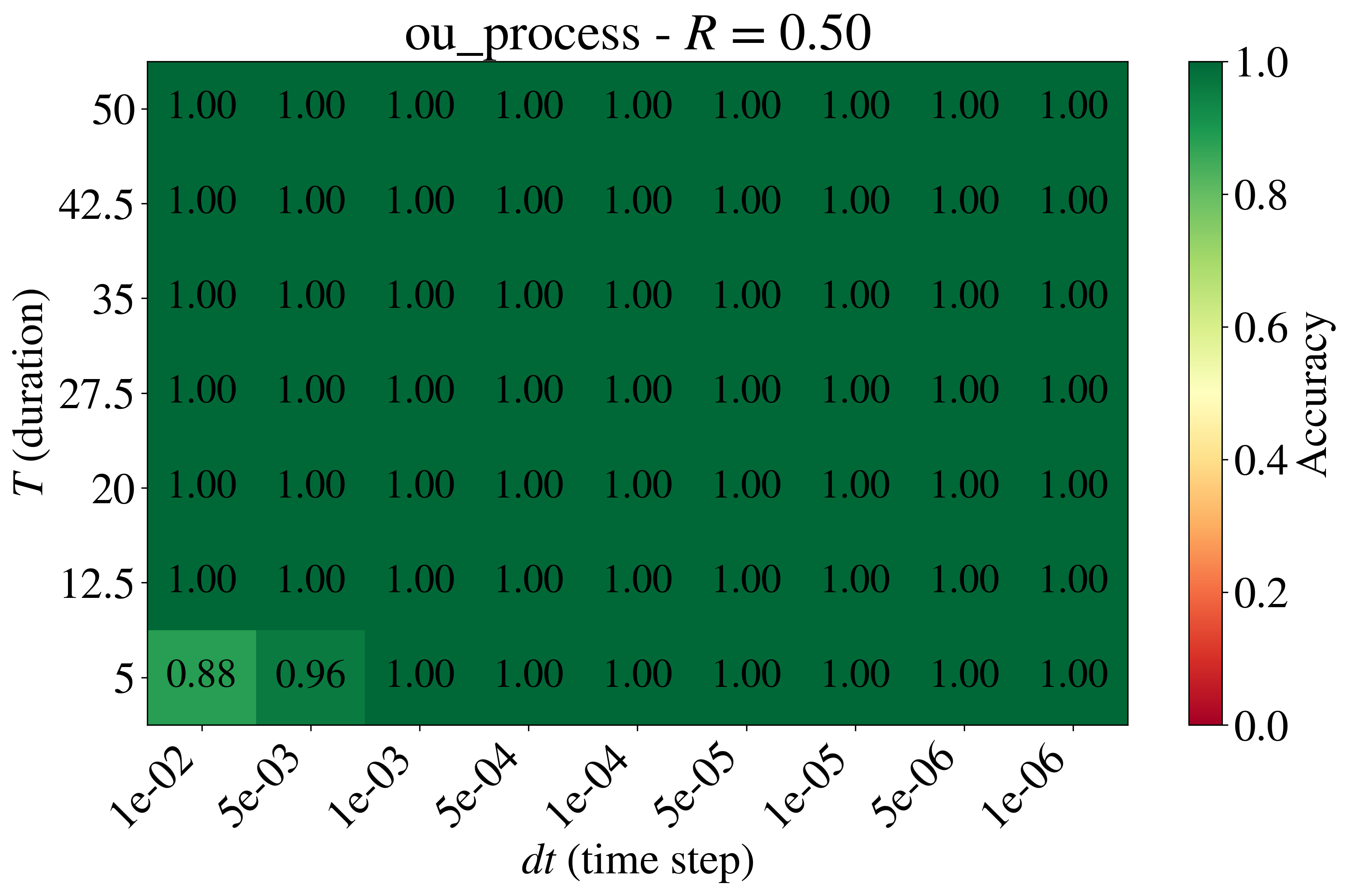}
\includegraphics[width=0.49\linewidth]{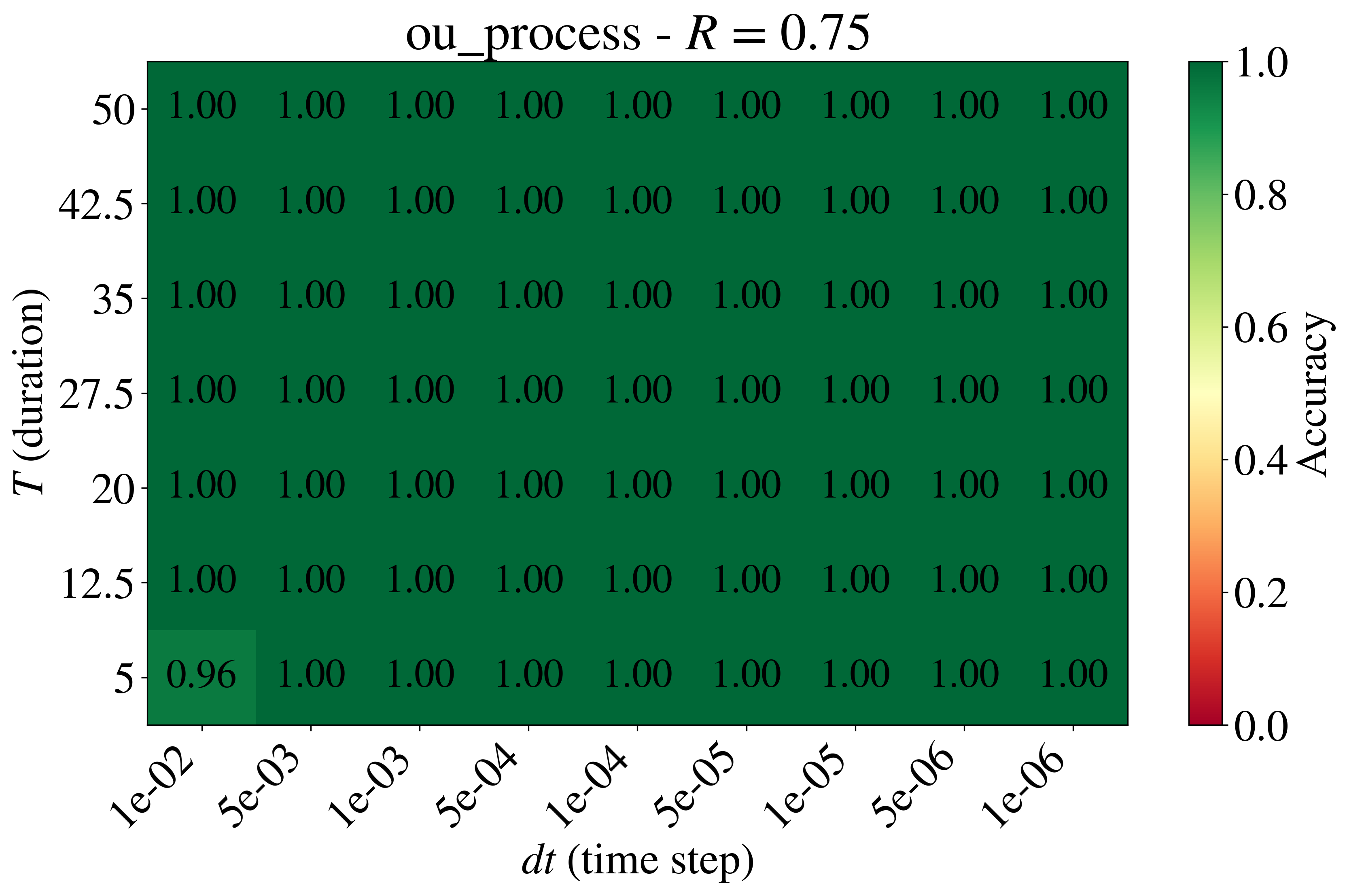}\includegraphics[width=0.49\linewidth]{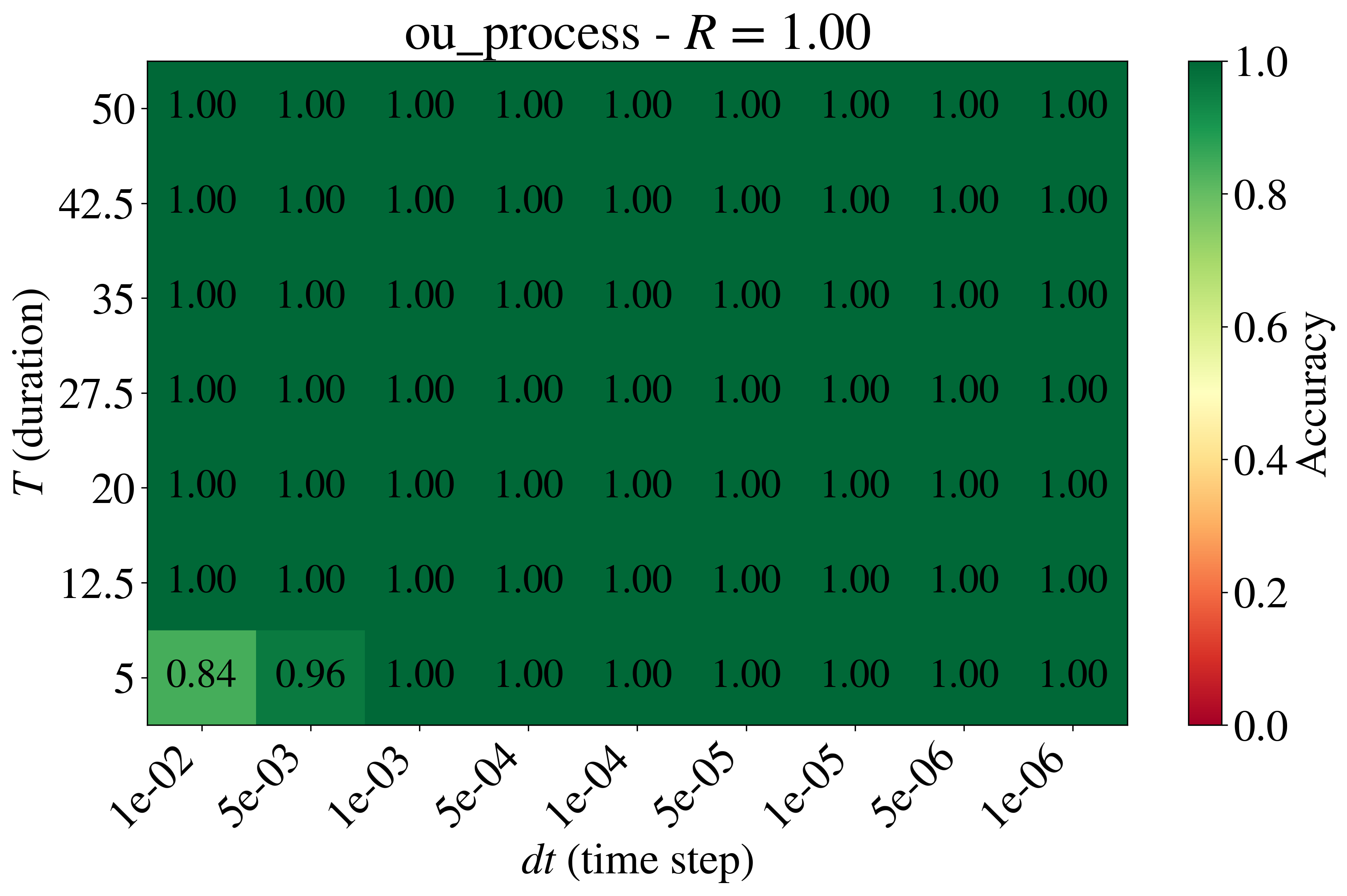}\caption{Heatmaps of accuracy for OU process, across various $R$.}
\label{fig:hmap_ou}
\end{figure}

\begin{figure}[!htbp]
\centering
\includegraphics[width=0.49\linewidth]{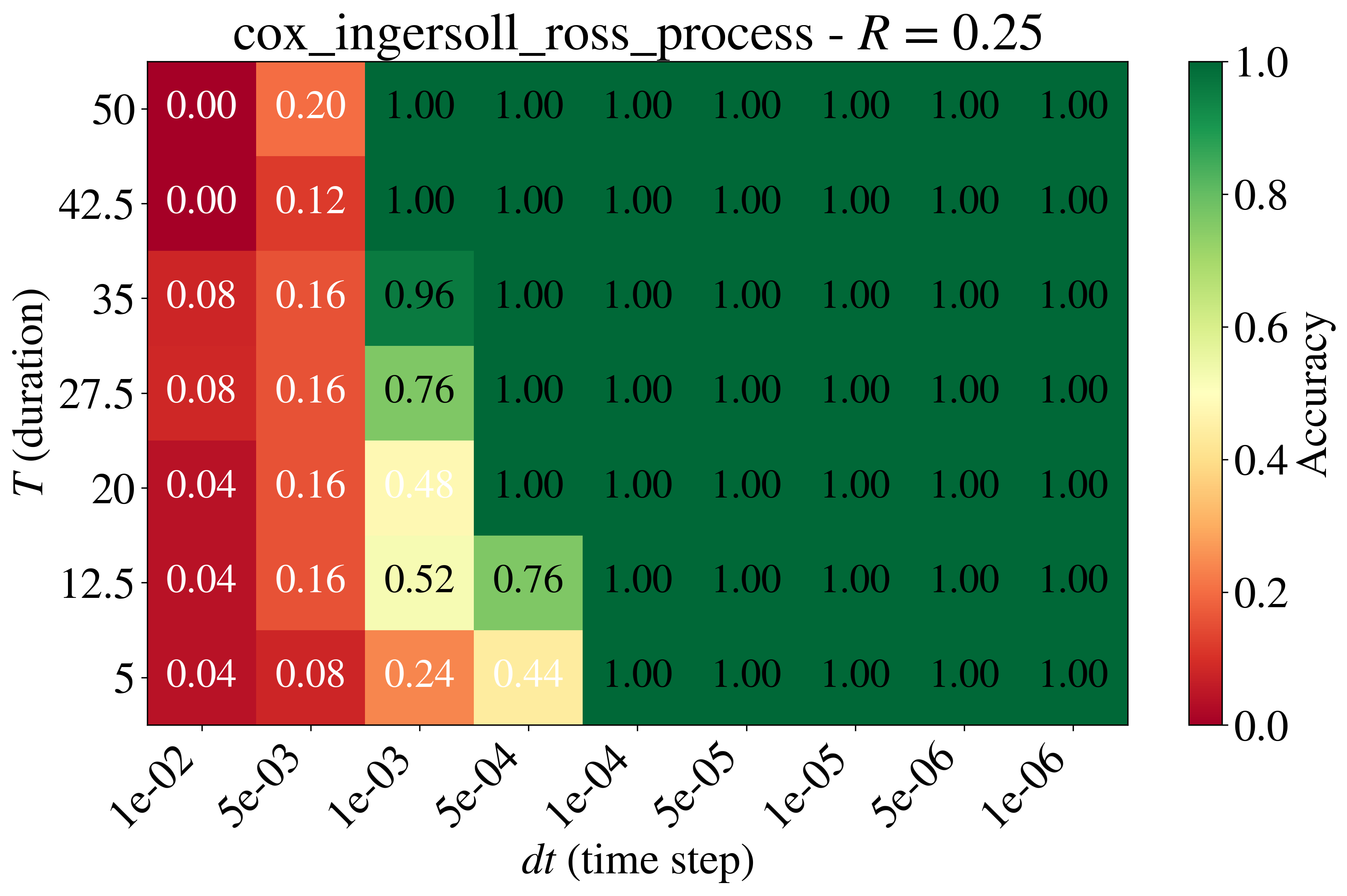}
\includegraphics[width=0.49\linewidth]{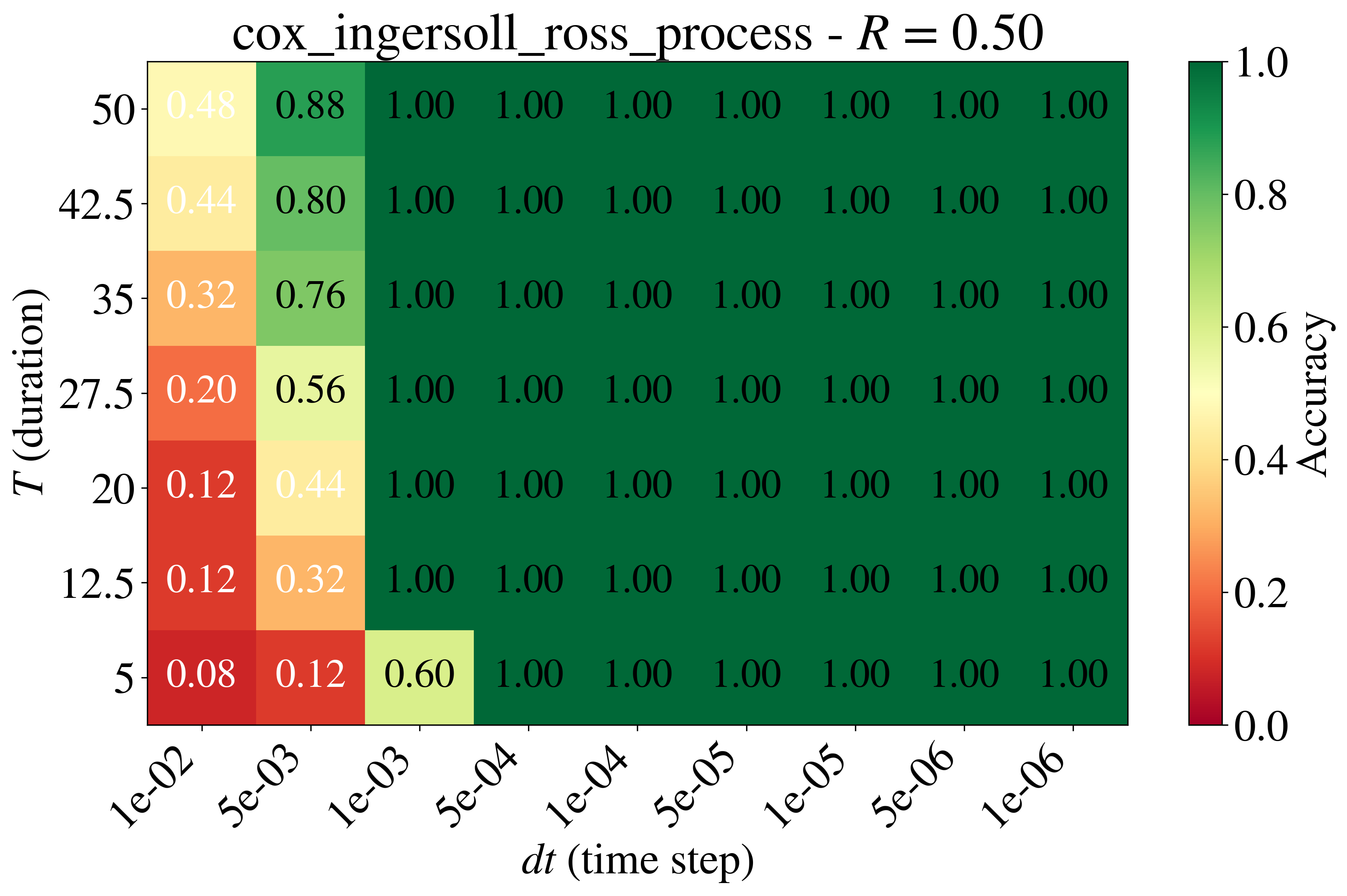}
\includegraphics[width=0.49\linewidth]{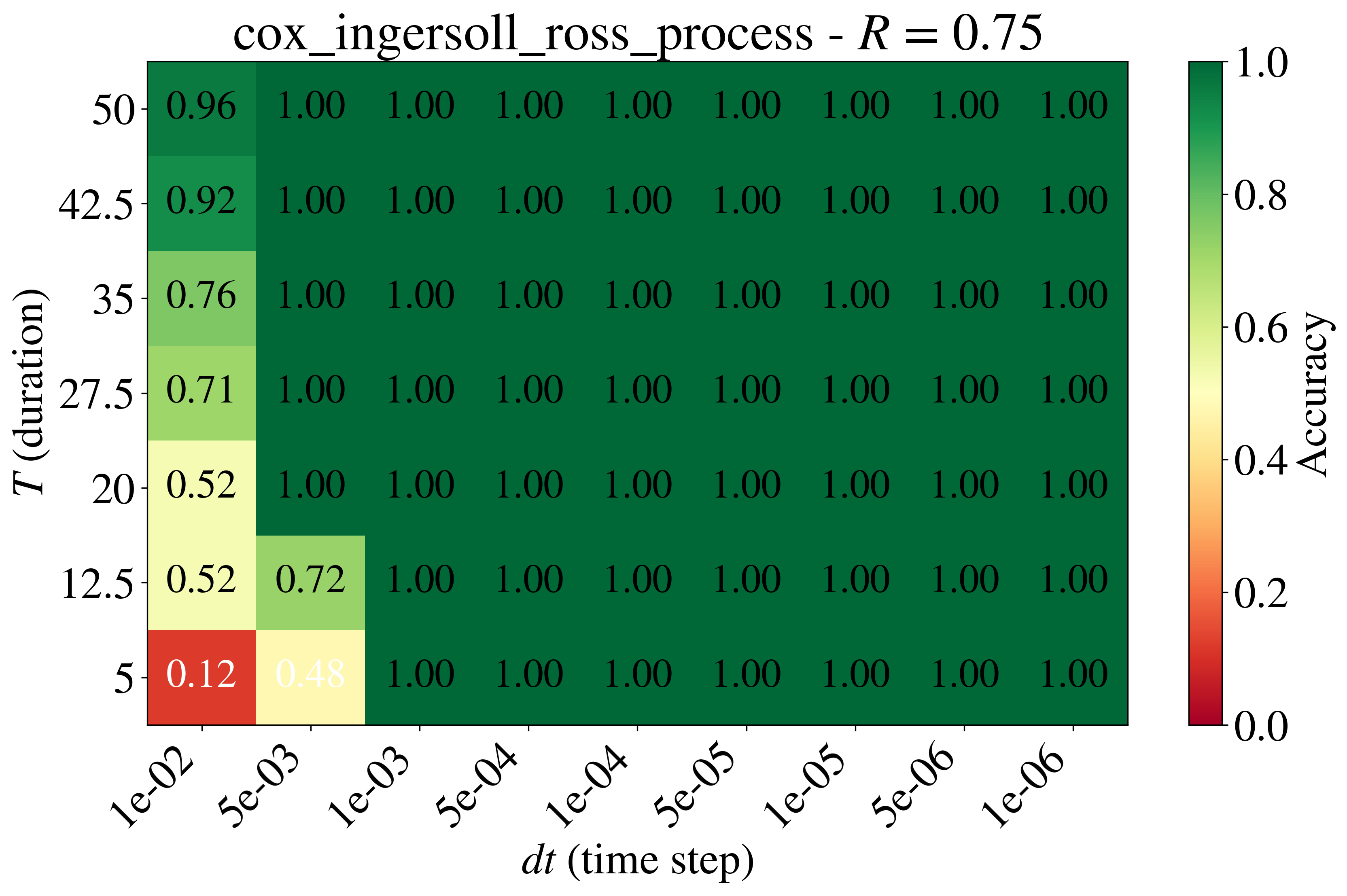}\includegraphics[width=0.49\linewidth]{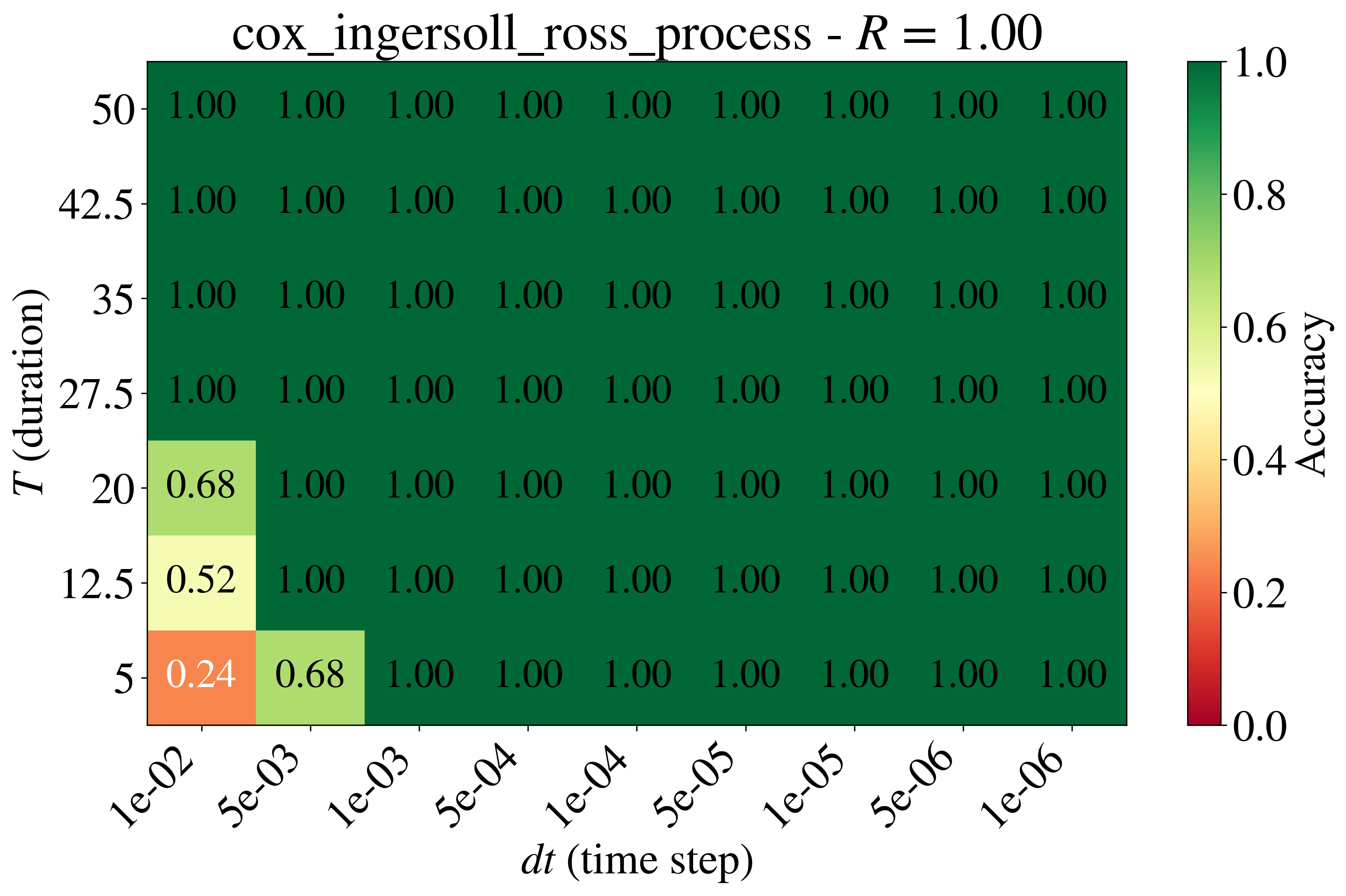}
\caption{Heatmaps of accuracy for CIR process, across various $R$.}
\label{fig:hmap_cir}
\end{figure}

We begin with classical diffusion processes, including the Ornstein-Uhlenbeck (OU) process, and the Cox-Ingersoll-Ross (CIR) model. Across all tested parameter regimes, these systems are classified as diffusion-like with near-perfect accuracy given sufficient discretization and sample length.
Fig.~\ref{fig:hmap_ou} and~\ref{fig:hmap_cir} show the heatmaps of accuracy for each of these systems. For OU processes, the classification accuracy is largely 100\%. For CIR, we see the accuracy increasing with \textit{R}, discretization $dt$ and signal length $T$.

\subsection{Deterministic Periodic System}

\begin{figure}[!htbp]
\centering
\includegraphics[width=0.49\linewidth]{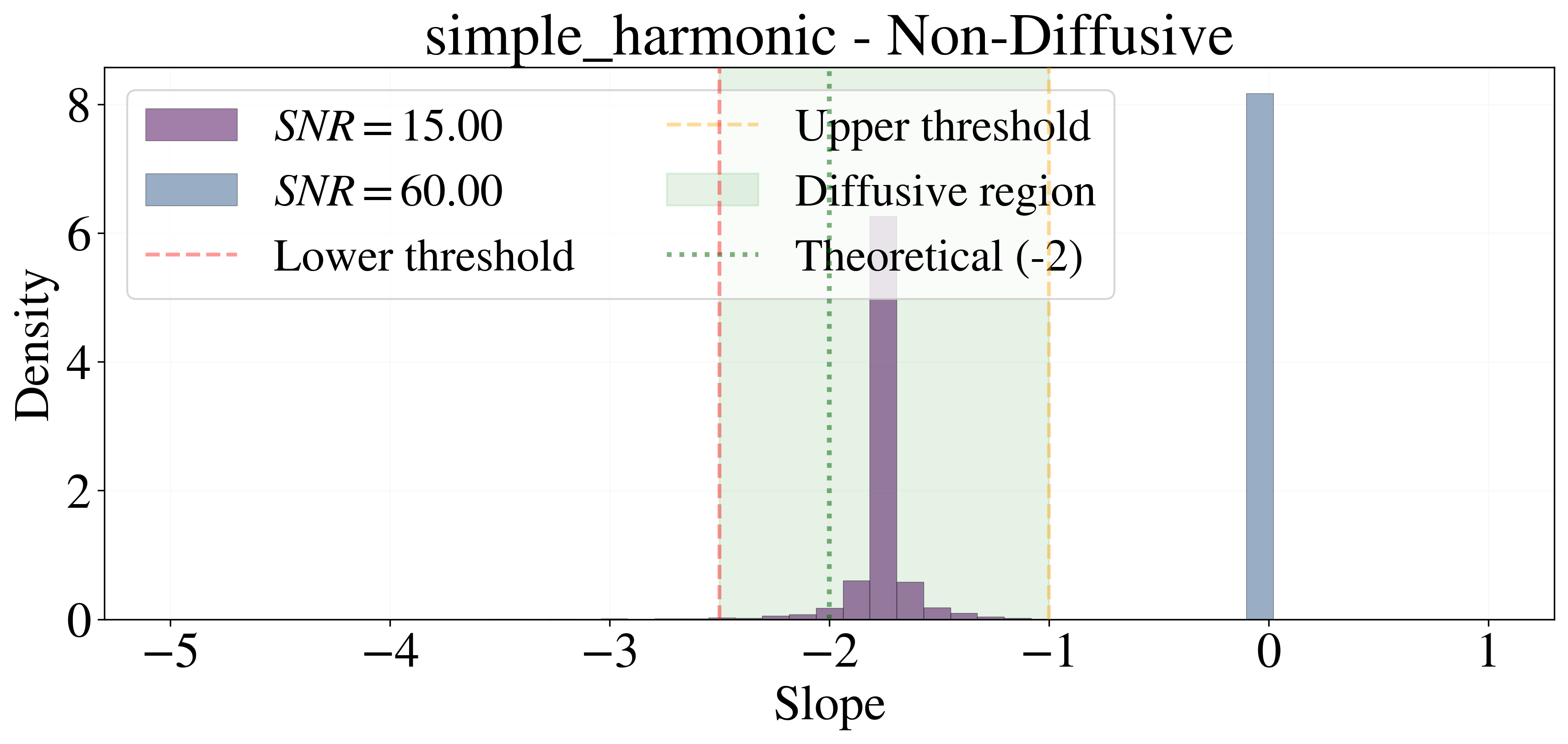}
\includegraphics[width=0.49\linewidth]{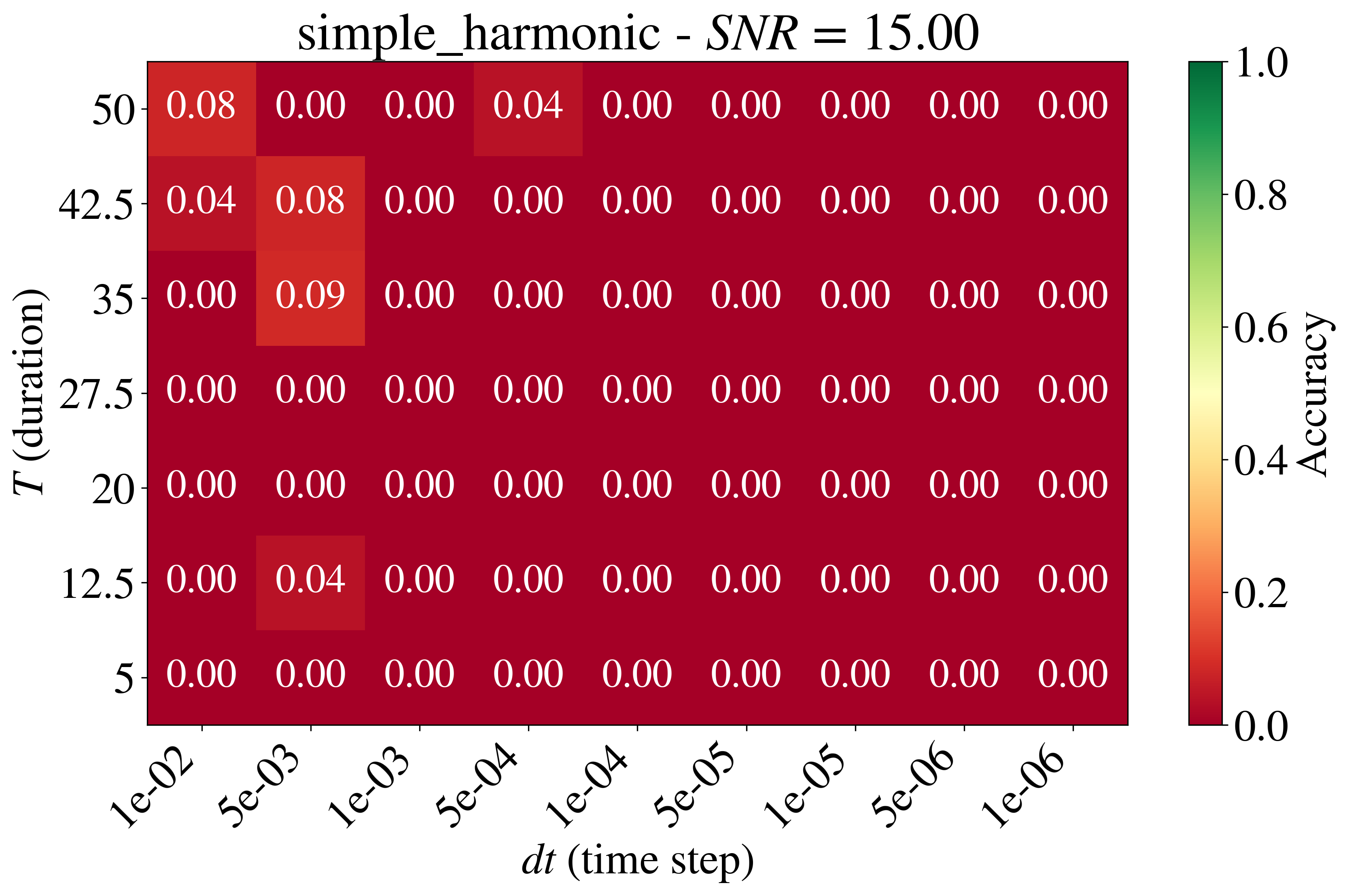}
\includegraphics[width=0.49\linewidth]{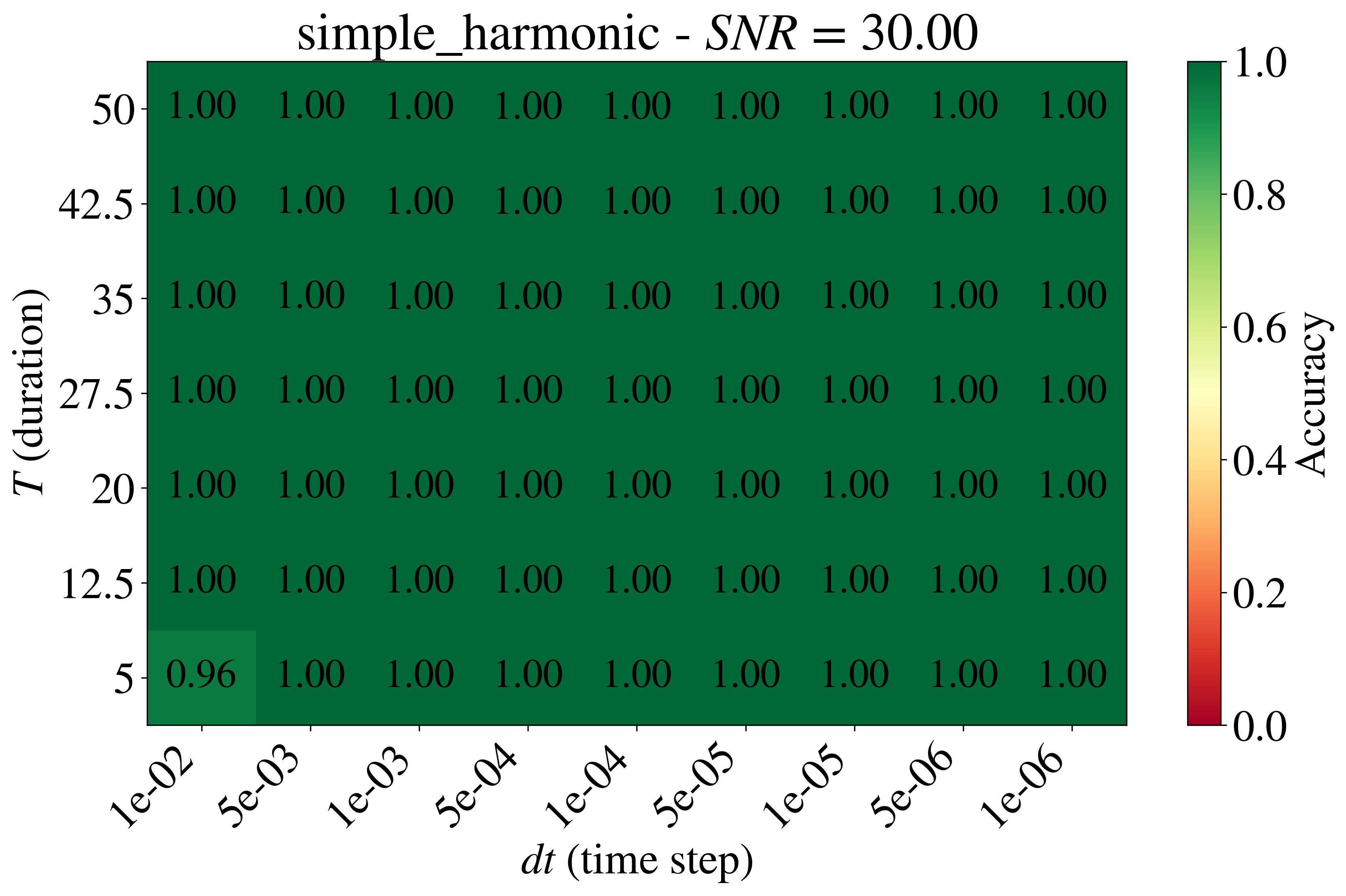}
\includegraphics[width=0.49\linewidth]{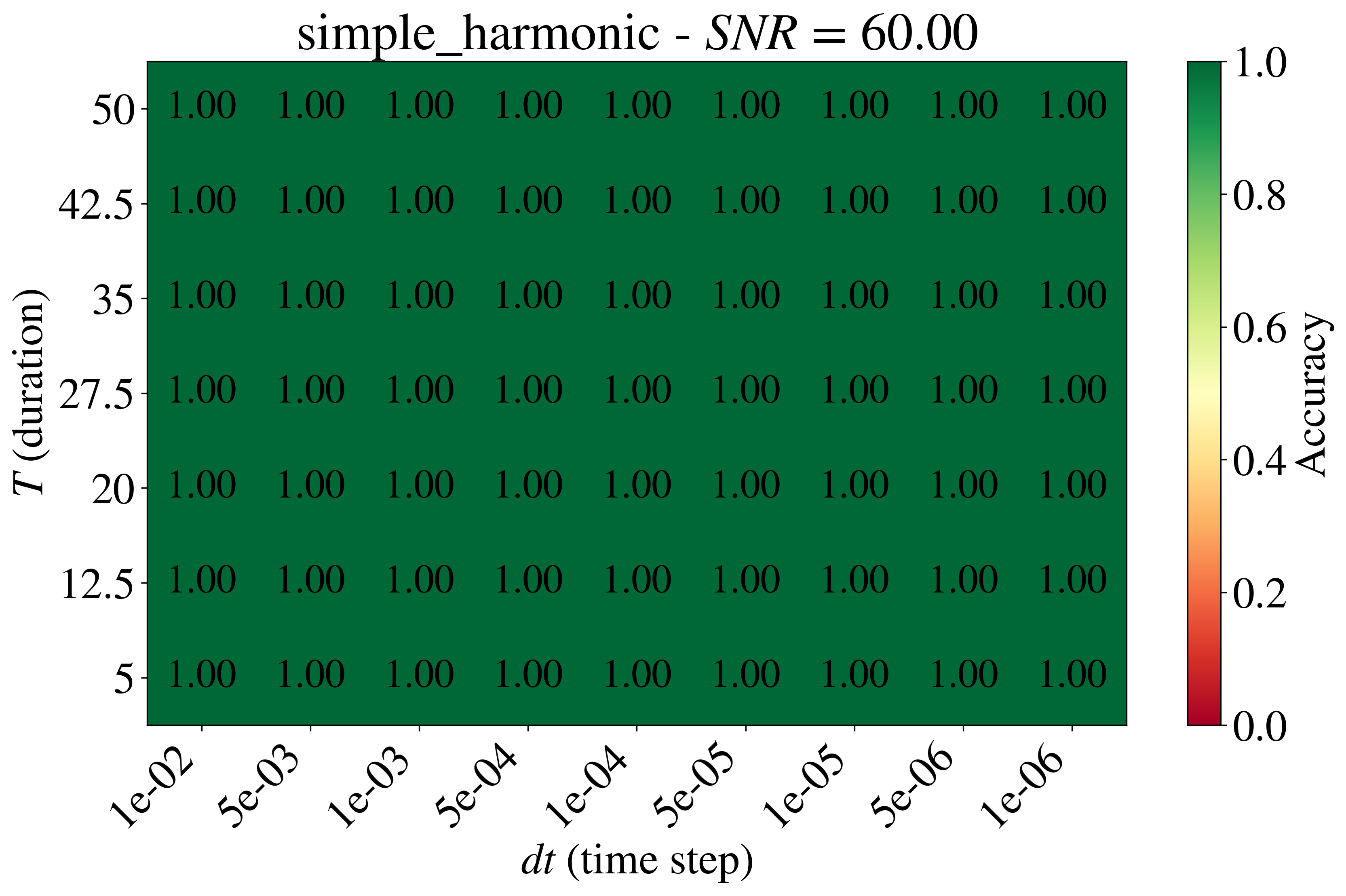}
\caption{Heatmaps of accuracy for simple harmonic motion, across various $SNR$.}
\label{fig:hmap_shm}
\end{figure}

Purely deterministic oscillators, such as the simple harmonic oscillator, are consistently classified as non-diffusive at low and moderate noise levels. In these regimes, excursion counts grow sub-quadratically in $\varepsilon^{-1}$, yielding slope estimates substantially larger than $-2$ and systematic deviations from the quadratic-variation-based prediction. As additive noise increases, these systems exhibit a gradual transition toward diffusion-like behavior. At low signal-to-noise ratios (typically SNR $\le 15$), the noise component dominates the local structure of the signal, and the excursion statistics approach those of a diffusion. This transition is visible in the excursion slope distributions and accuracy heatmaps plotted against noise level and sampling resolution, shown in Fig.~\ref{fig:hmap_shm}.

\subsection{Deterministic Chaotic Maps}

\begin{figure}[!htbp]
\centering
\includegraphics[width=0.49\linewidth]{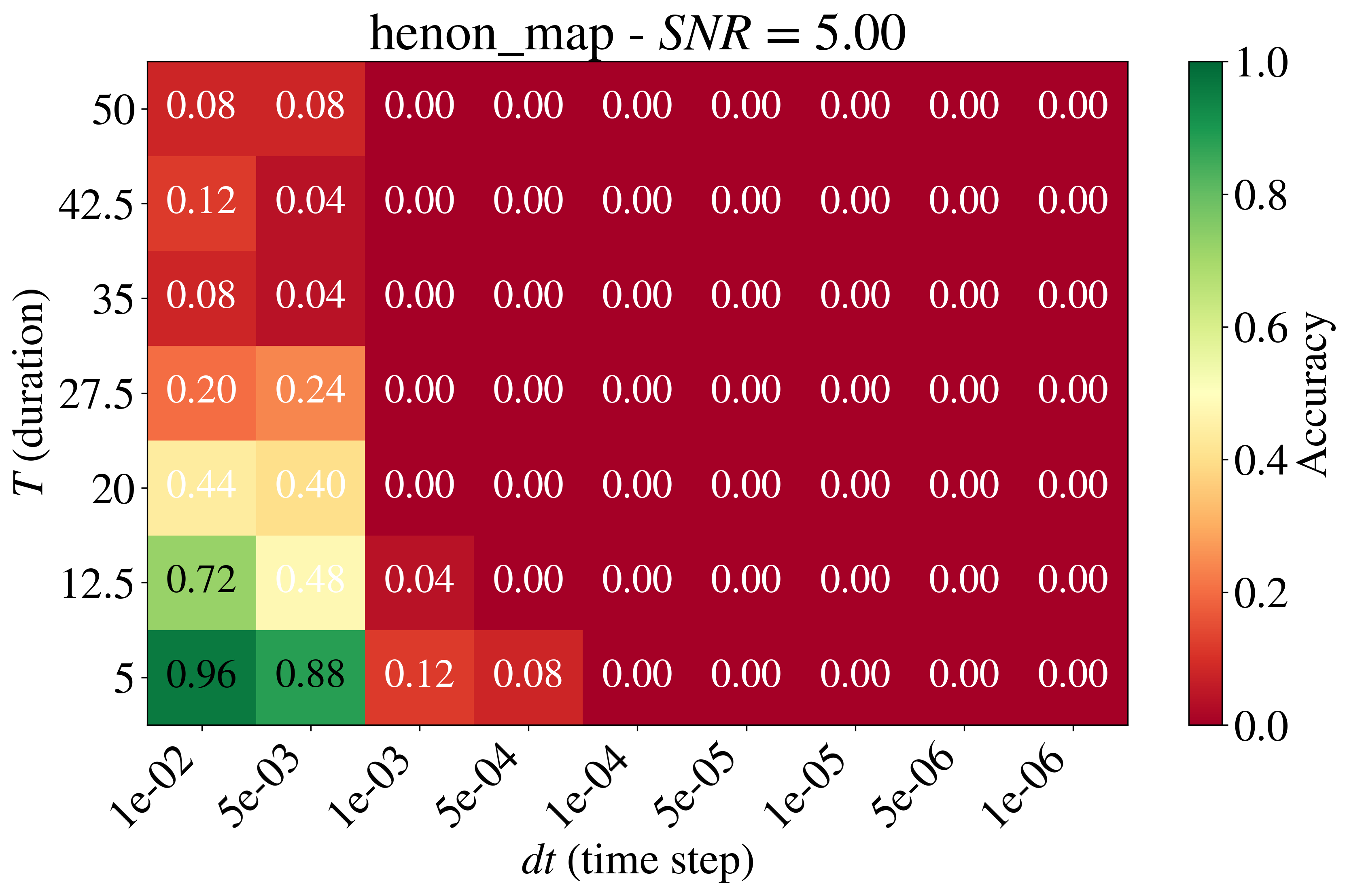}
\includegraphics[width=0.49\linewidth]{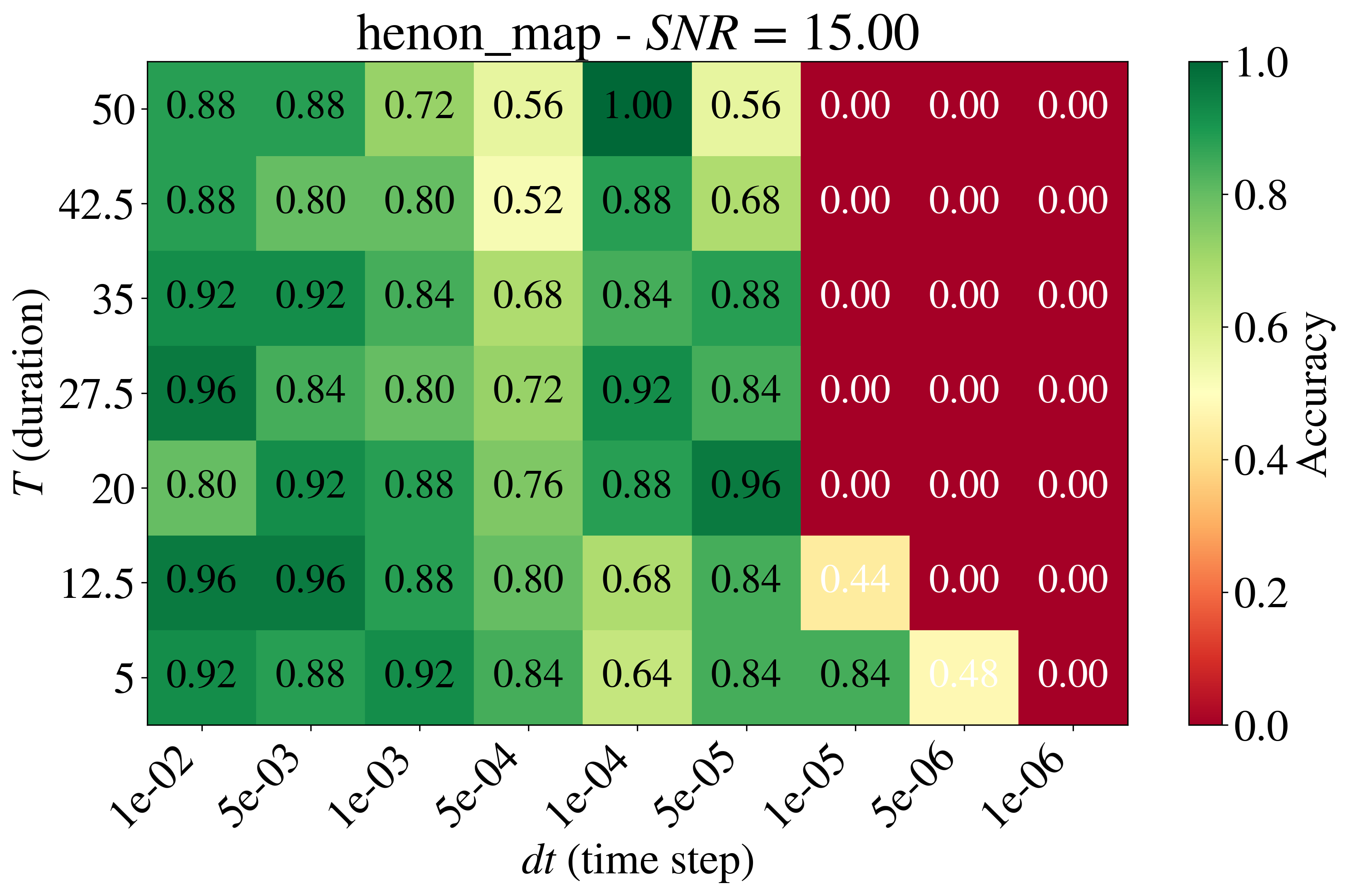}
\includegraphics[width=0.49\linewidth]{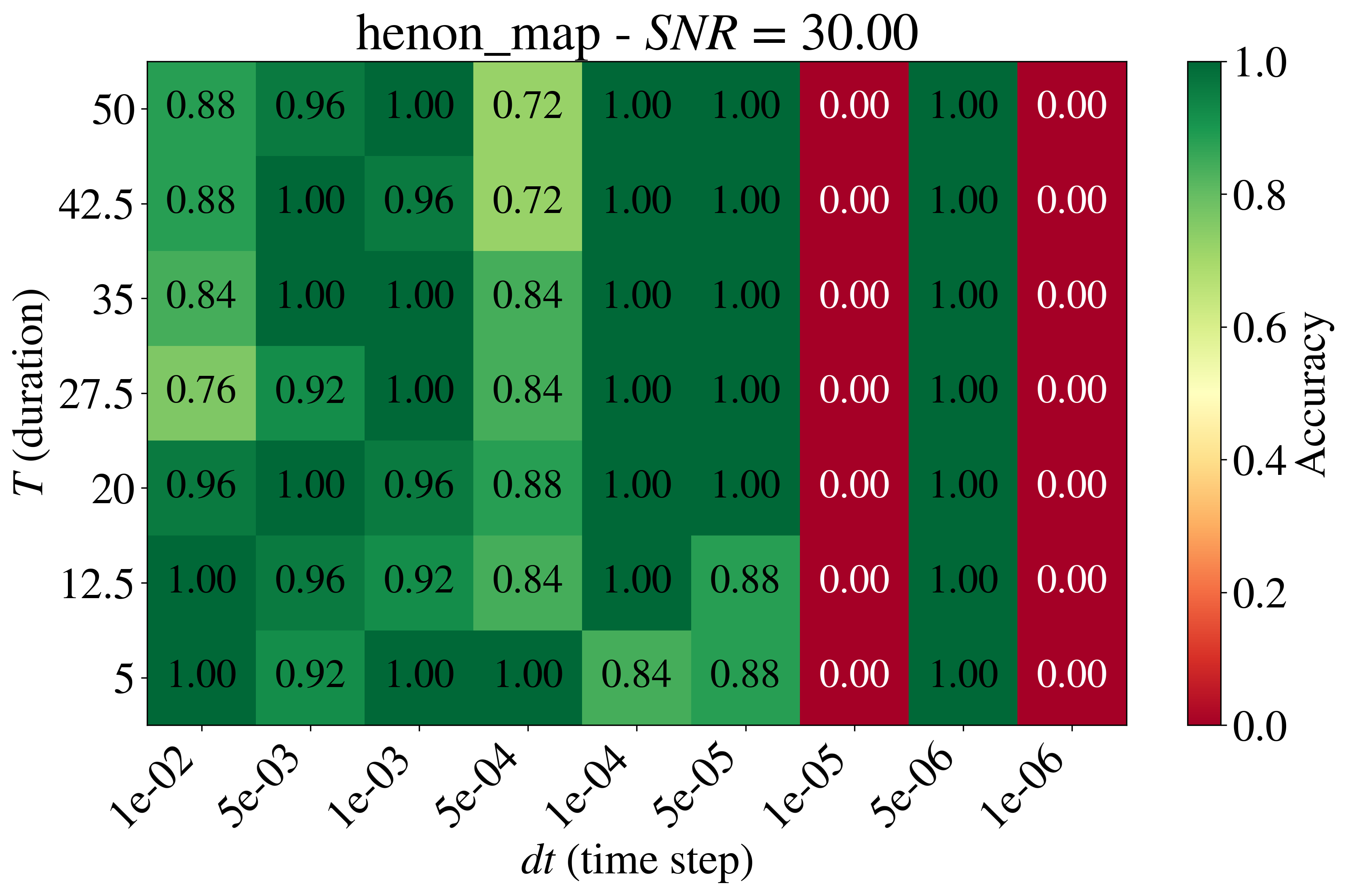}
\includegraphics[width=0.49\linewidth]{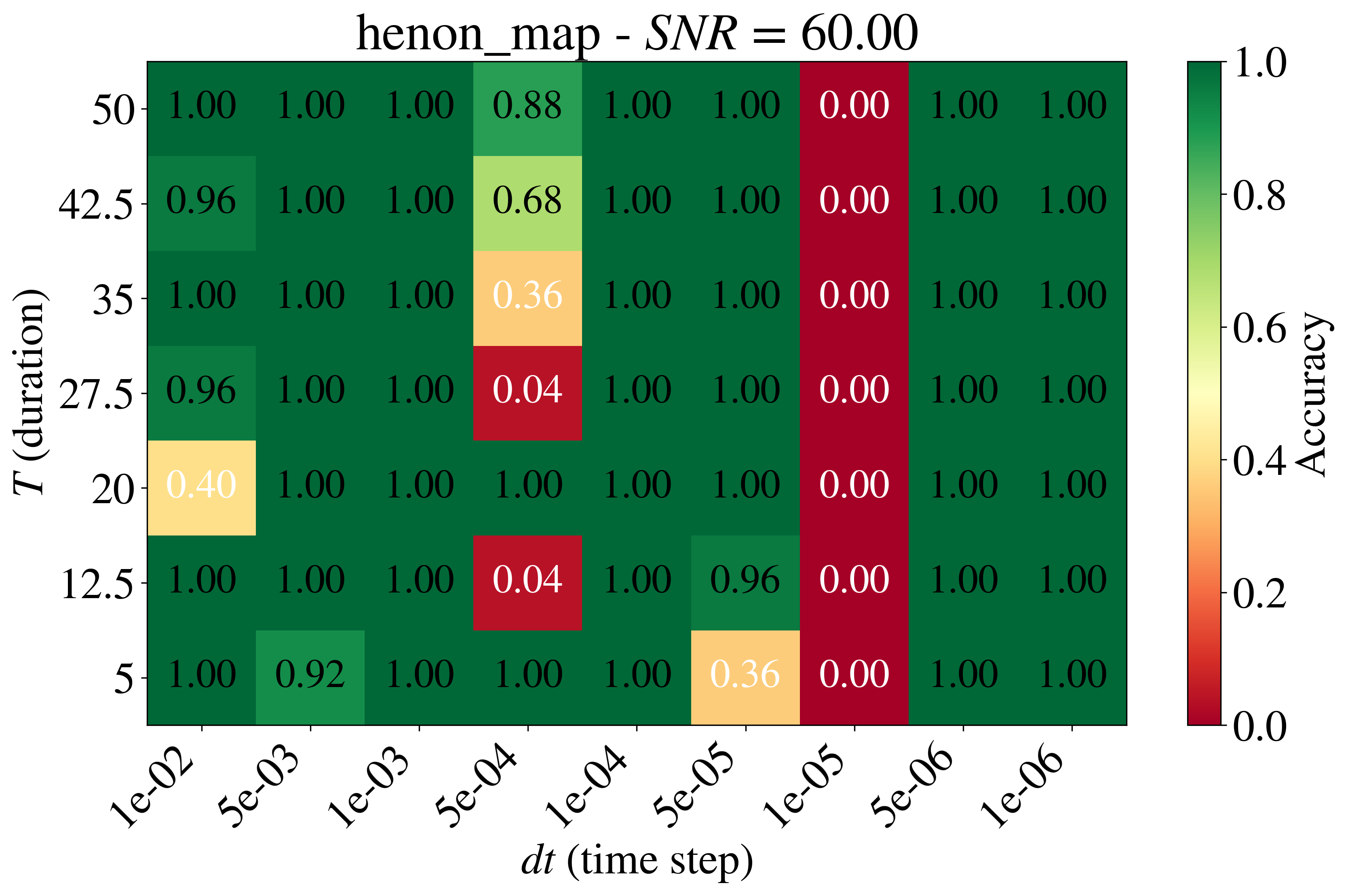}
\caption{Heatmaps of accuracy for Henon map, across various $SNR$.}
\label{fig:hmap_henon}
\end{figure}

\begin{figure}[!htbp]
\centering
\includegraphics[width=0.49\linewidth]{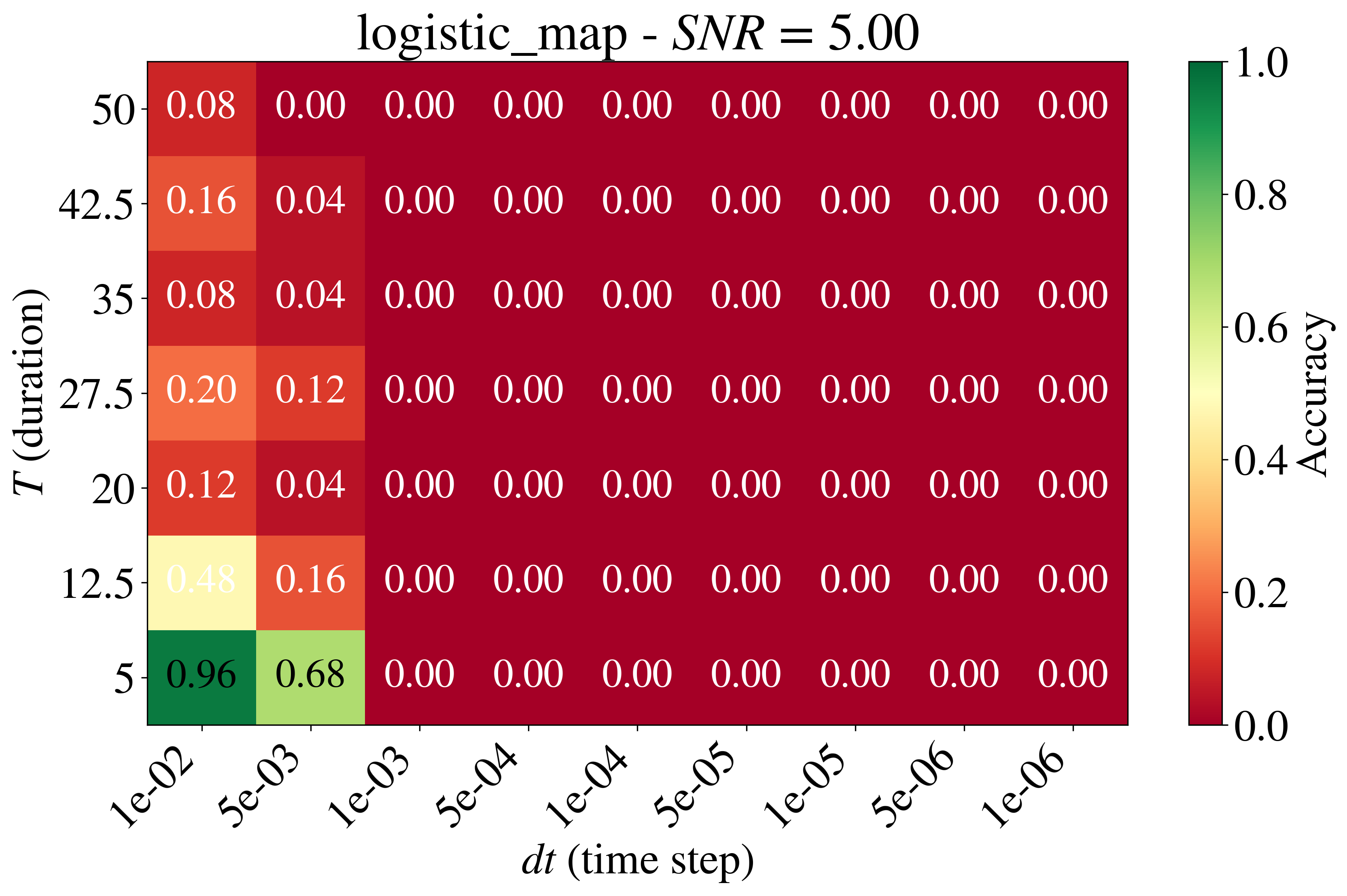}
\includegraphics[width=0.49\linewidth]{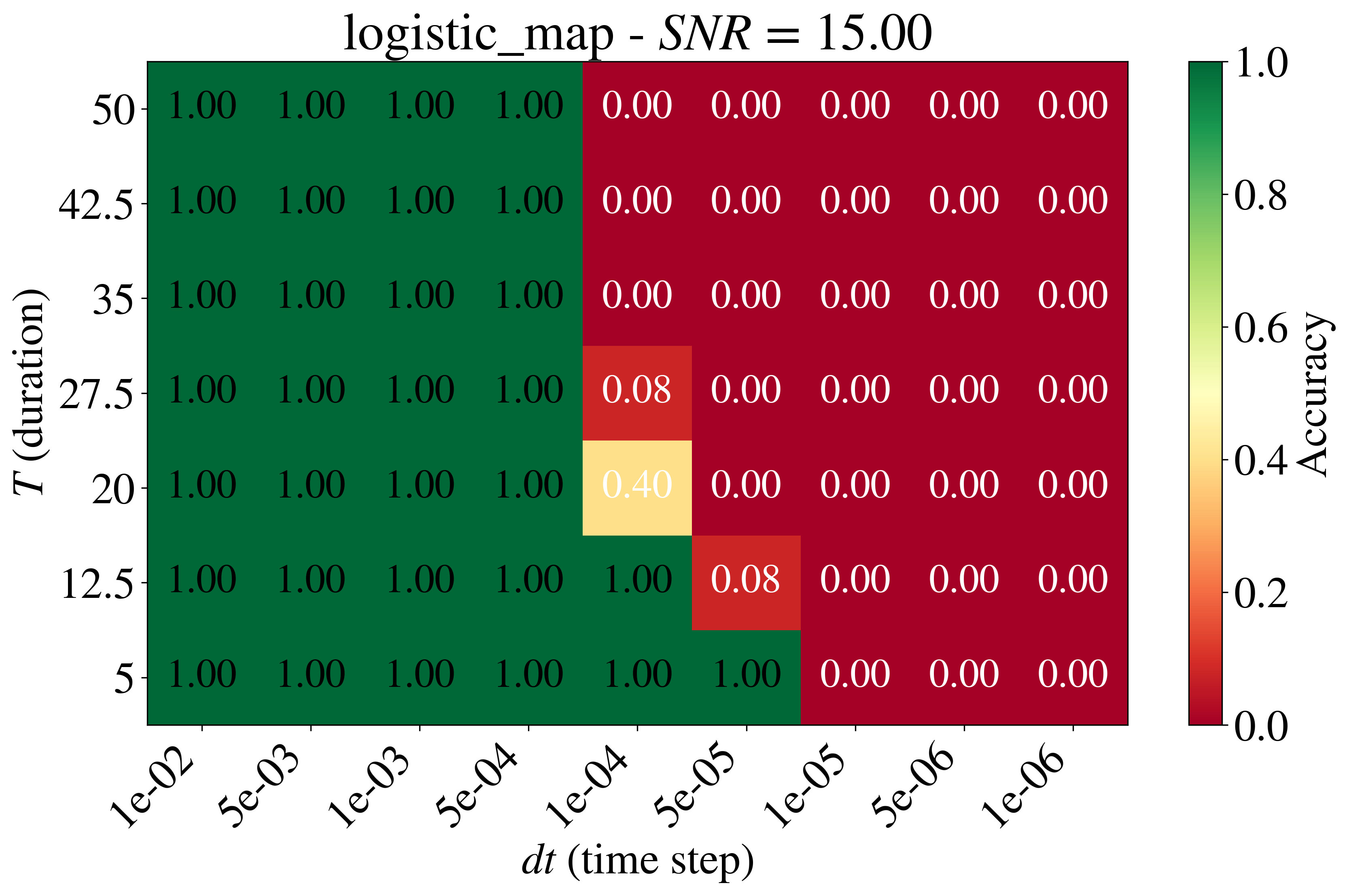}
\includegraphics[width=0.49\linewidth]{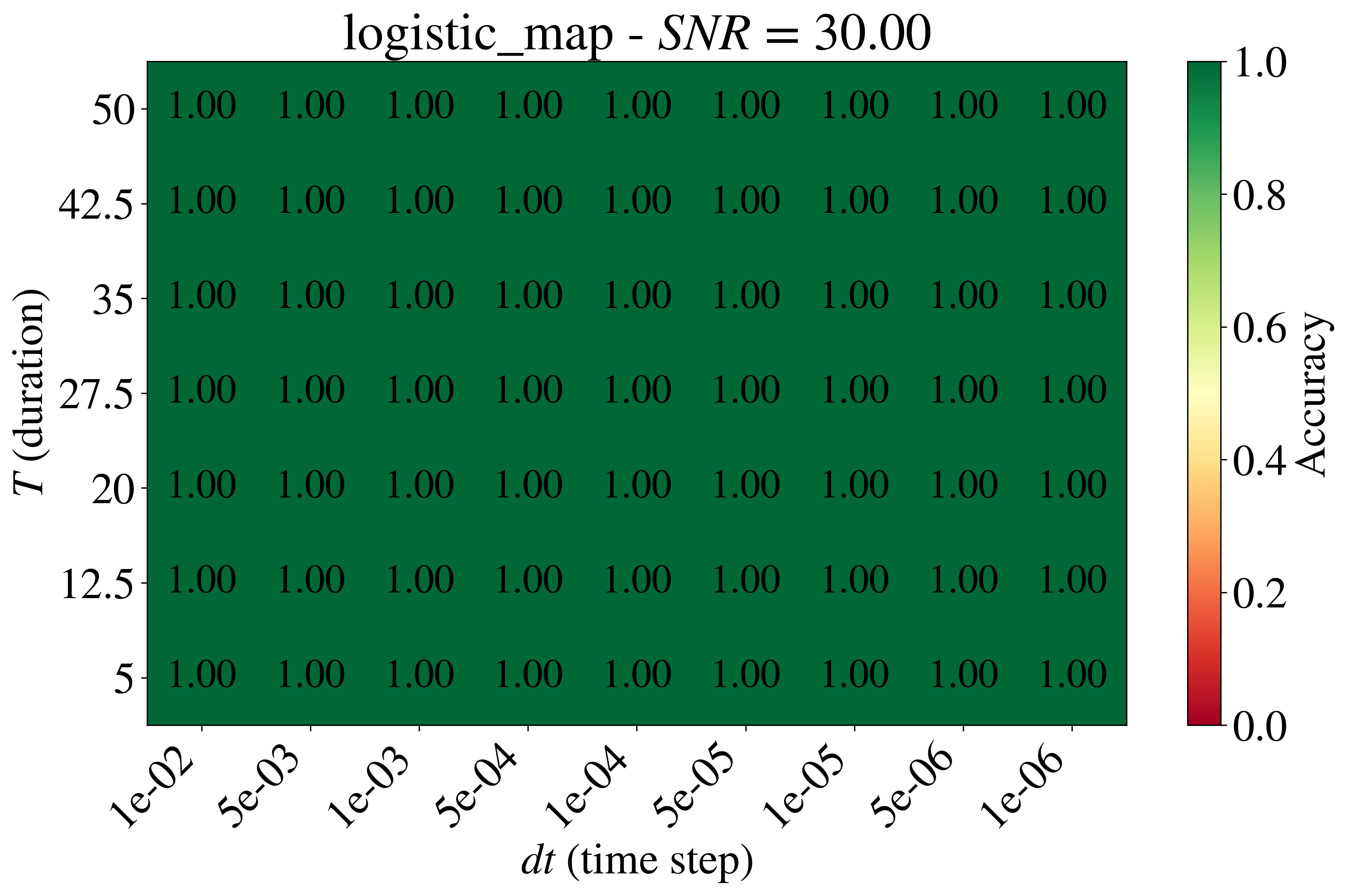}
\includegraphics[width=0.49\linewidth]{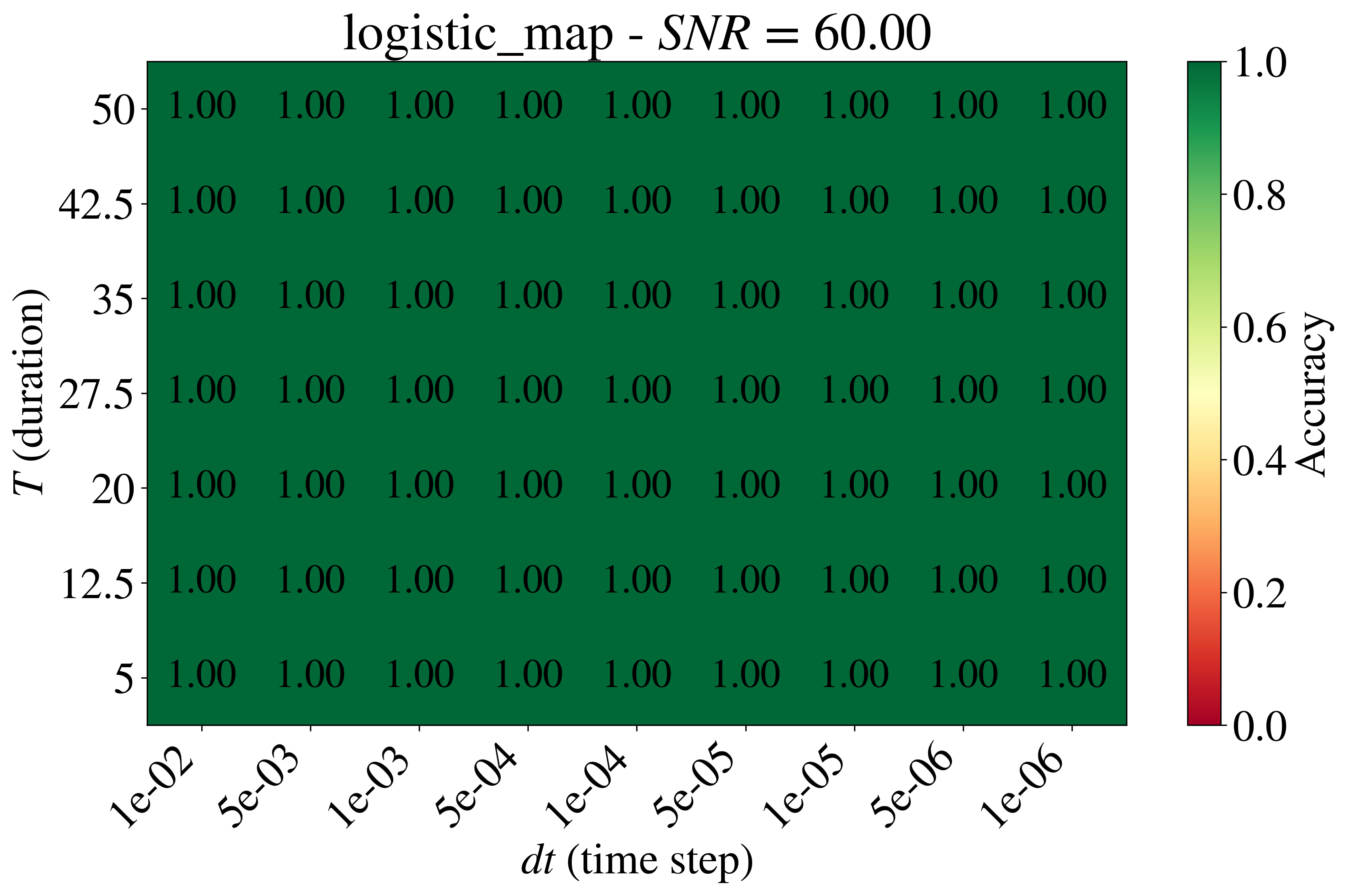}
\caption{Heatmaps of accuracy for logistic map, across various $SNR$.}
\label{fig:hmap_logistic}
\end{figure}

\begin{figure}[!htbp]
\centering
\includegraphics[width=0.49\linewidth]{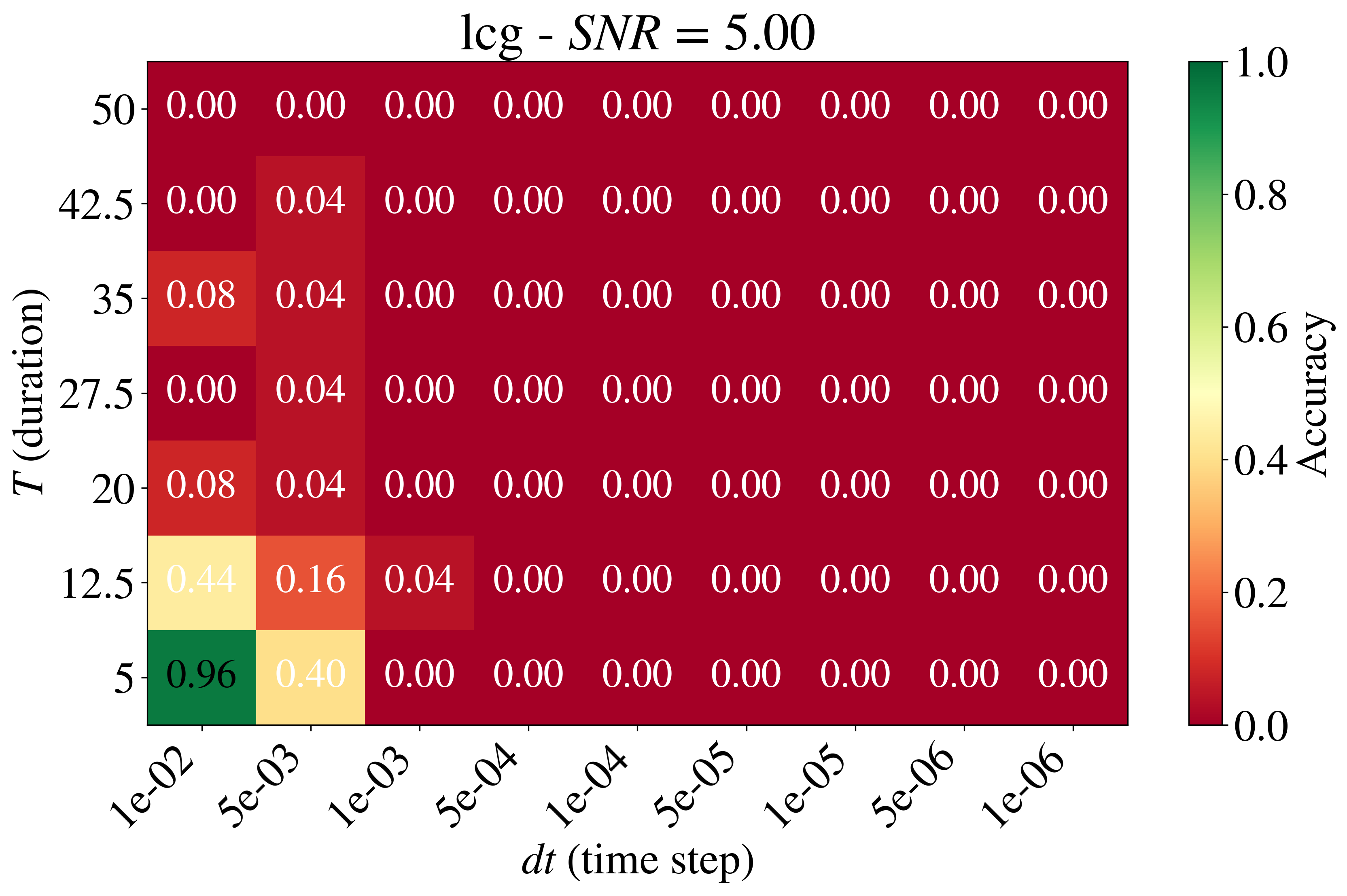}
\includegraphics[width=0.49\linewidth]{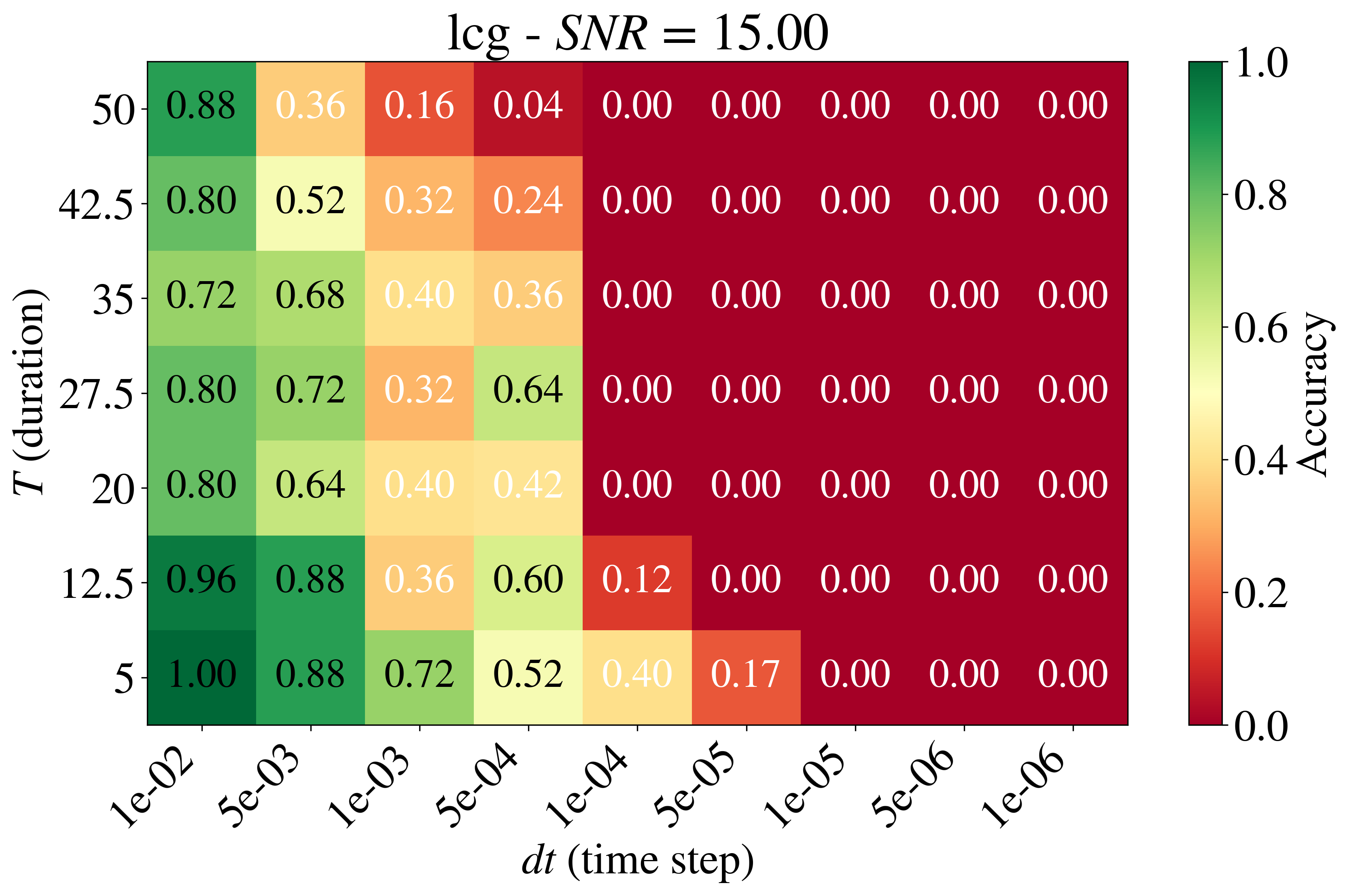}
\includegraphics[width=0.49\linewidth]{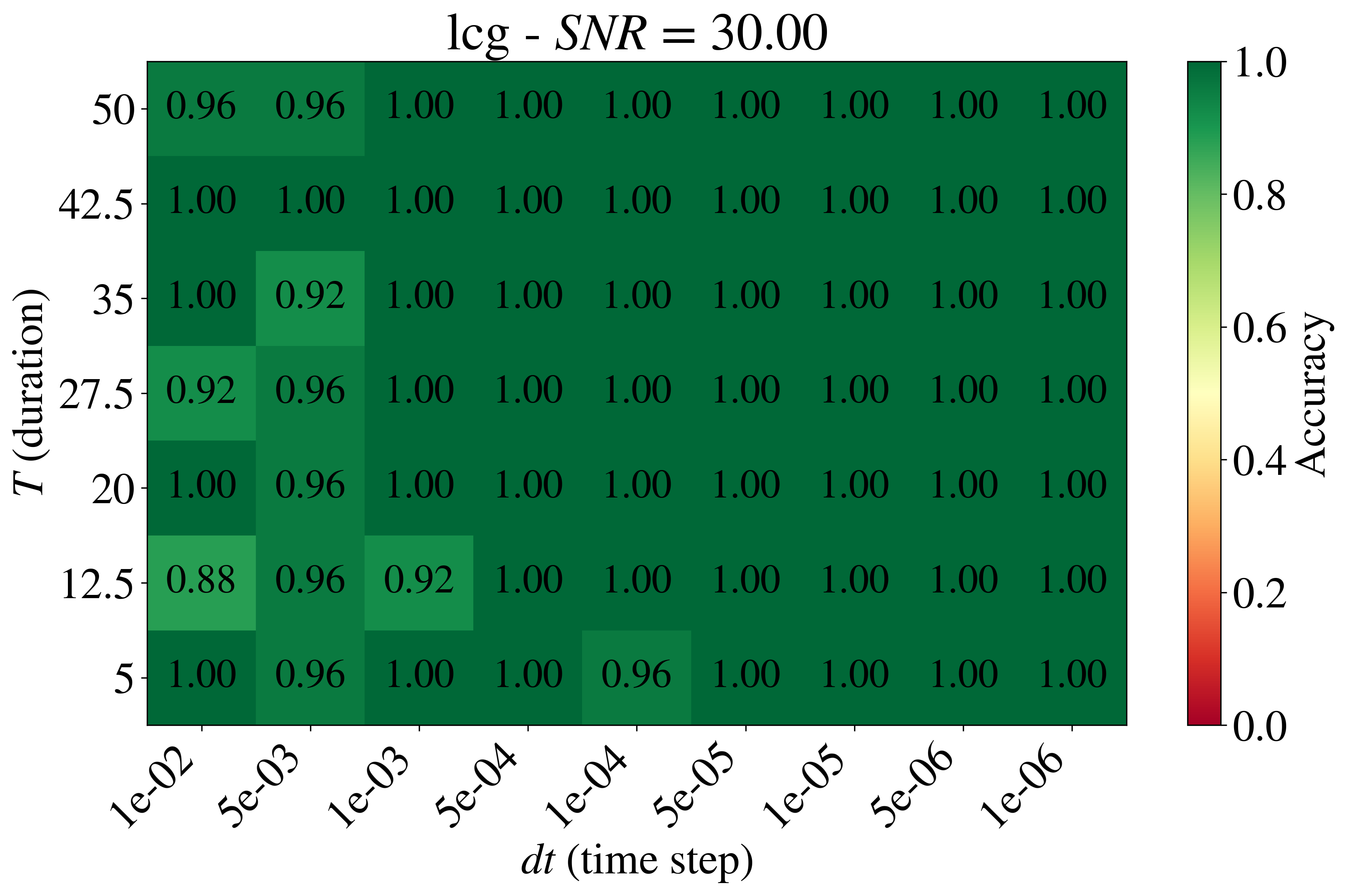}
\includegraphics[width=0.49\linewidth]{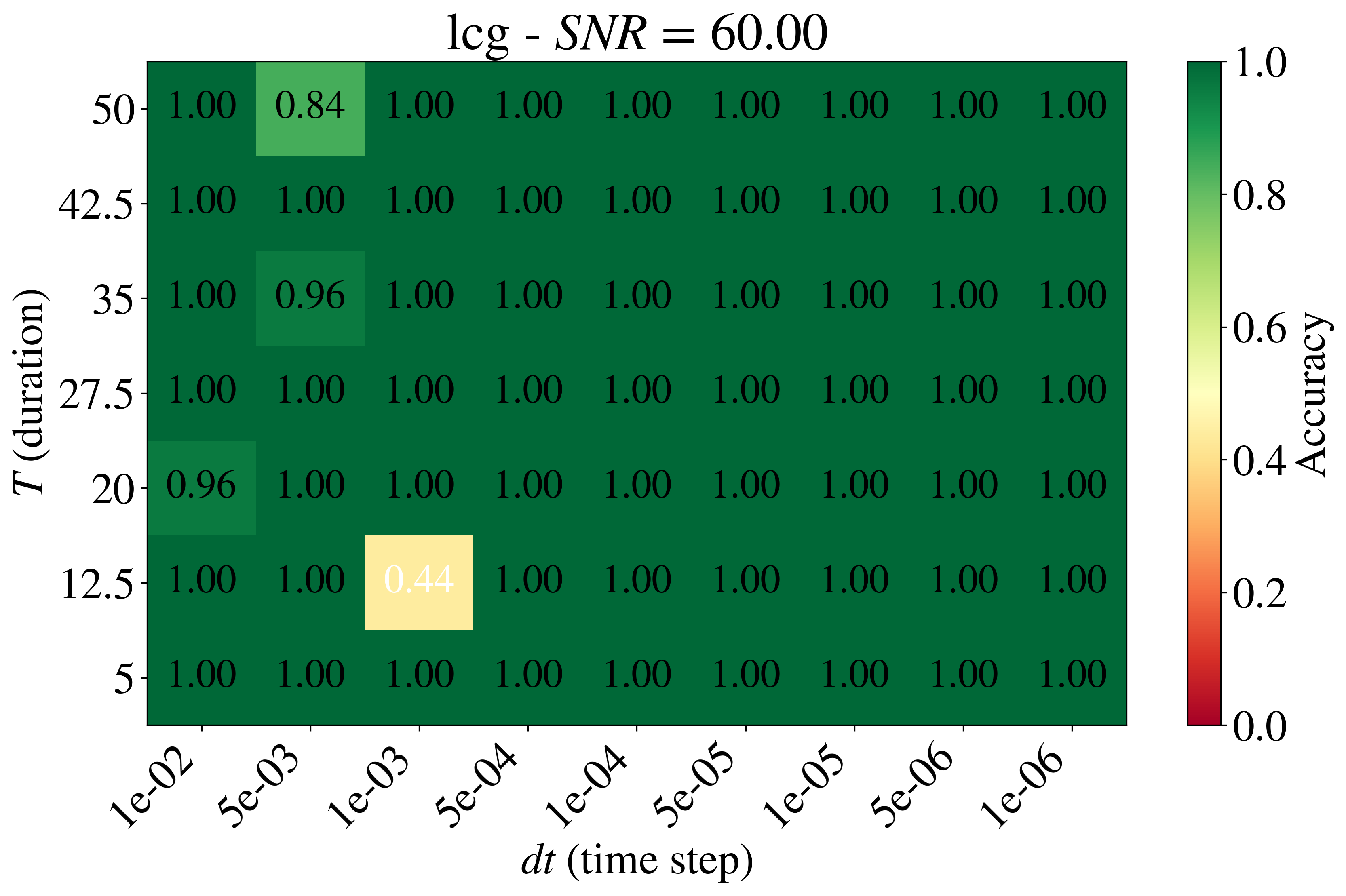}
\caption{Heatmaps of accuracy for LCG map, across various $SNR$.}
\label{fig:hmap_lcg}
\end{figure}

We next consider deterministic chaotic maps including Henon, logistic, and the notoriously difficult Linear Congruential Generator (LCG) maps. In the absence of noise, these systems are robustly classified as non-diffusive despite their apparent irregularity. Their excursion statistics fail to satisfy the $\varepsilon^{-2}$ scaling law, reflecting the absence of finite quadratic variation in the continuous--time limit. When additive noise is introduced, these systems display sharp transitions from deterministic-like to diffusion-like classification. Notably, it is interesting that for chaotic maps, the accuracy increases as $dt$ and $T$ decrease instead of the other way around. See Fig.~\ref{fig:hmap_henon}-~\ref{fig:hmap_lcg} for heatmaps of accuracy. 

\subsection{Deterministic Chaotic Systems}

\begin{figure}[!htbp]
\centering
\includegraphics[width=0.49\linewidth]{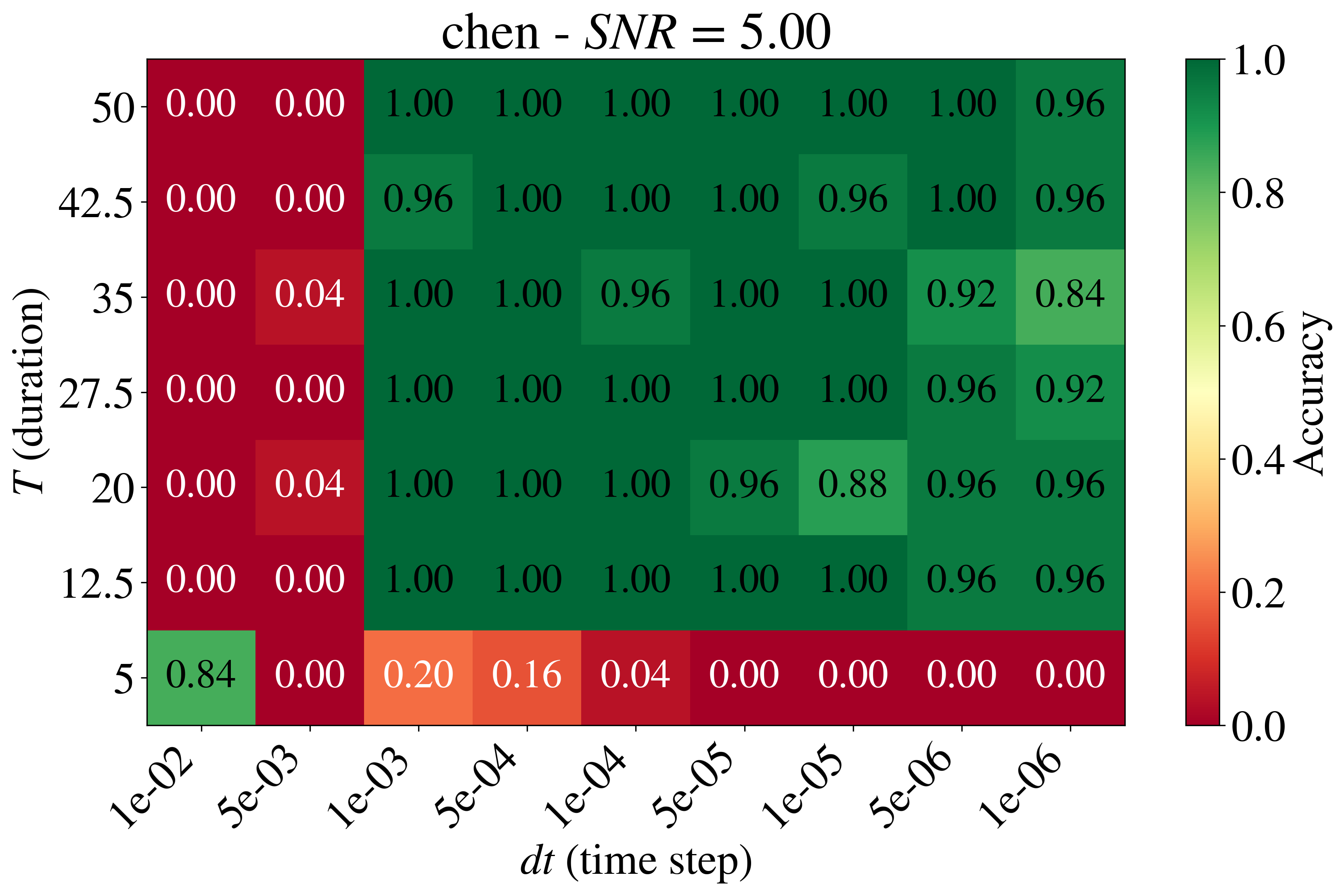}
\includegraphics[width=0.49\linewidth]{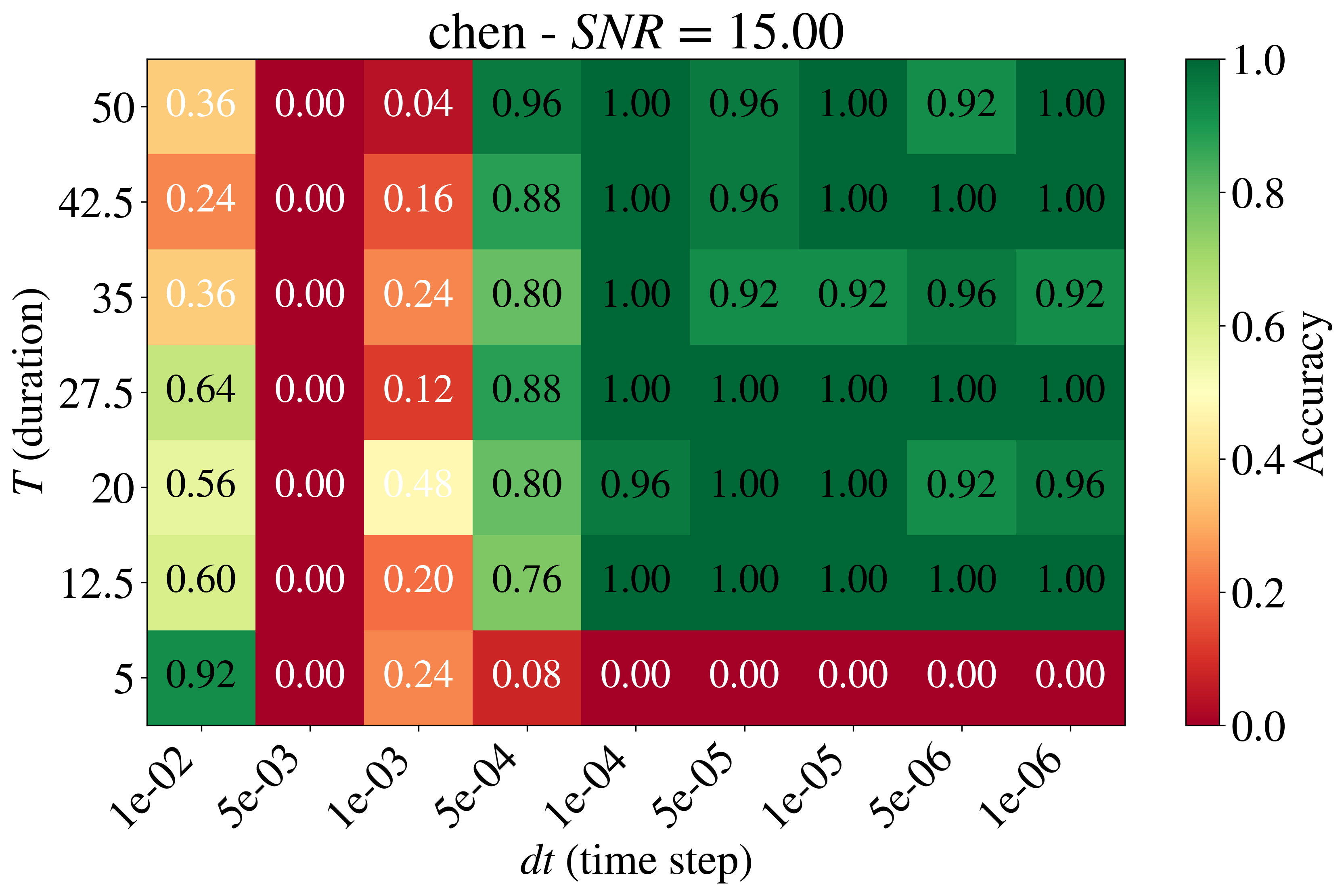}
\includegraphics[width=0.49\linewidth]{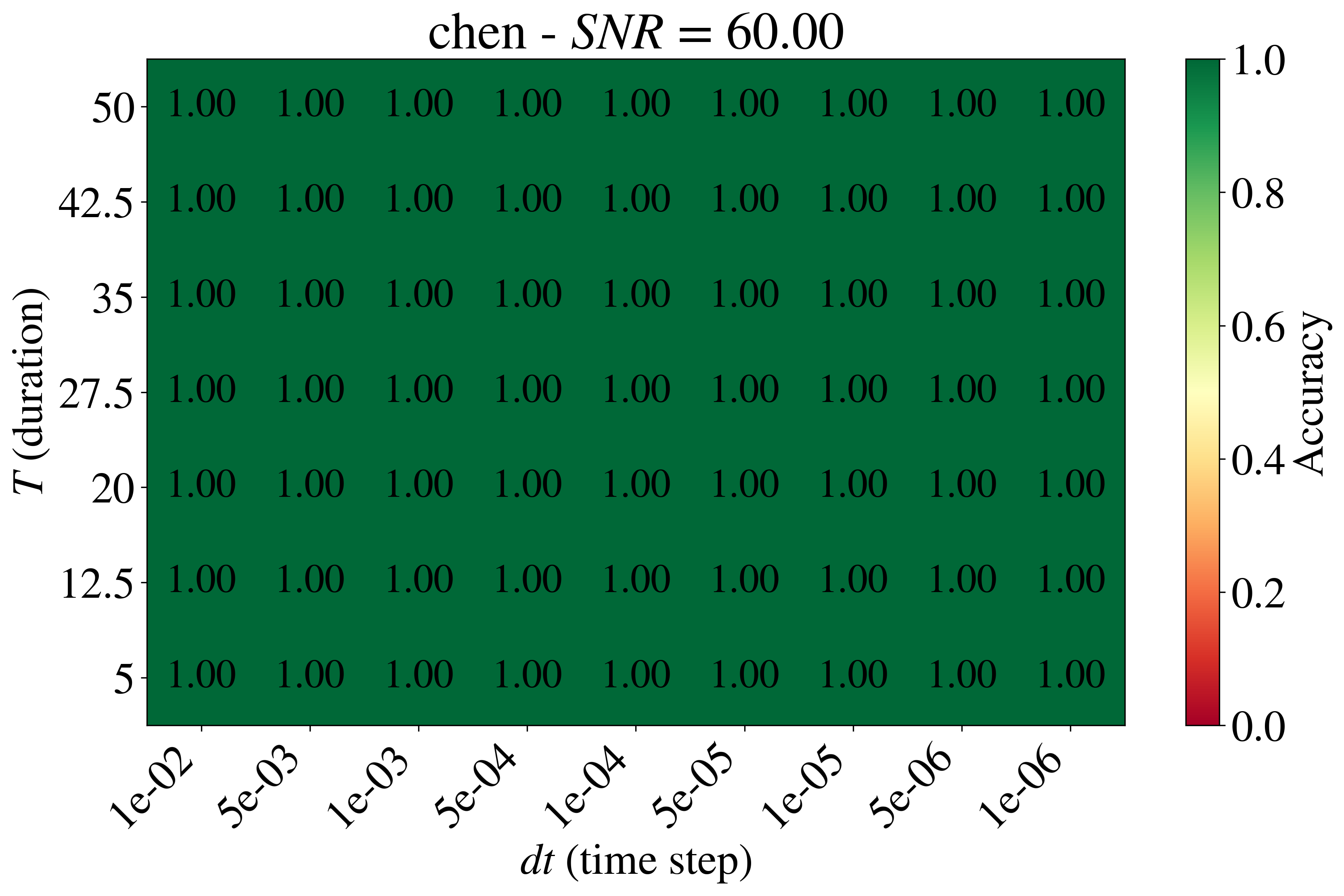}
\includegraphics[width=0.49\linewidth]{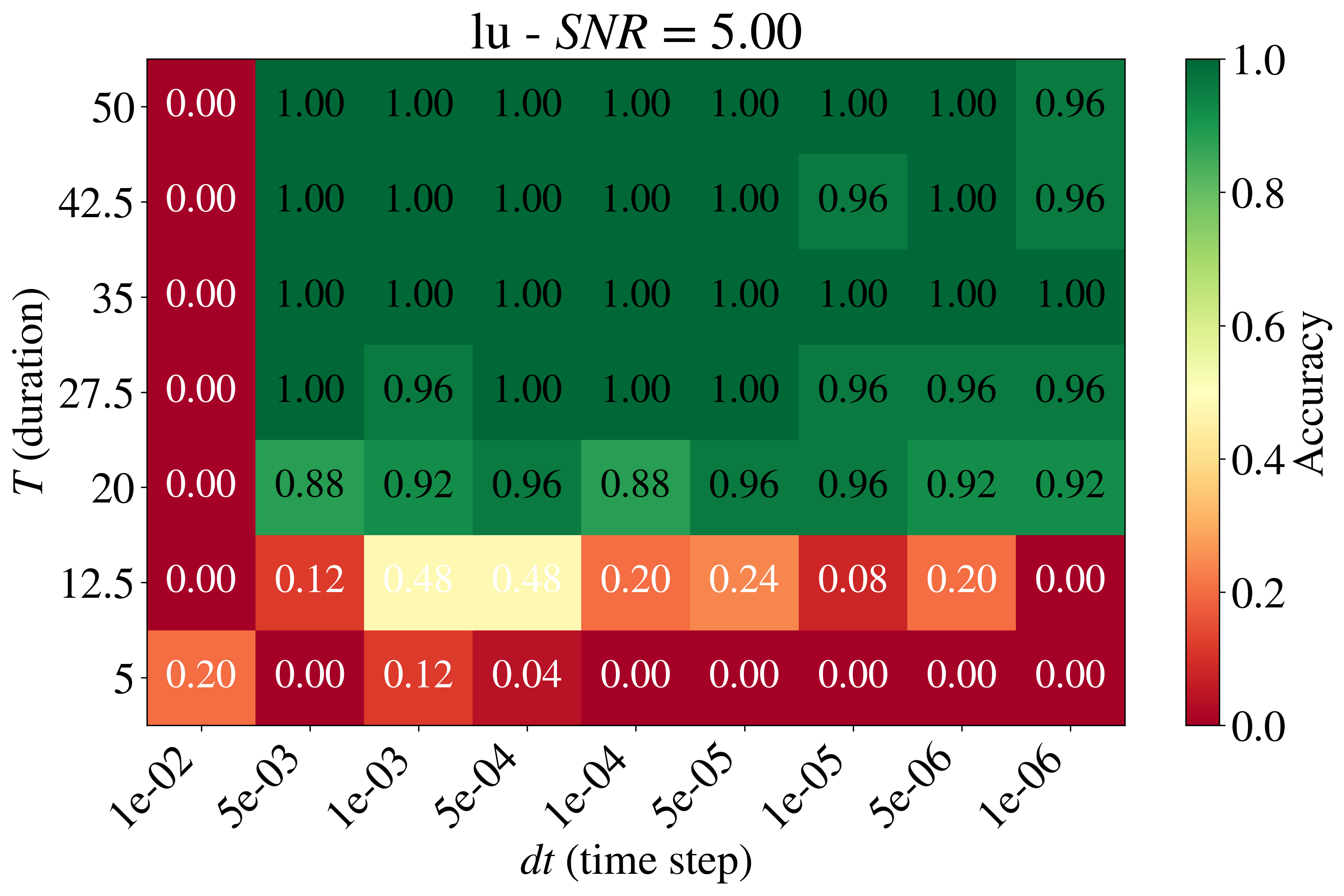}
\includegraphics[width=0.49\linewidth]{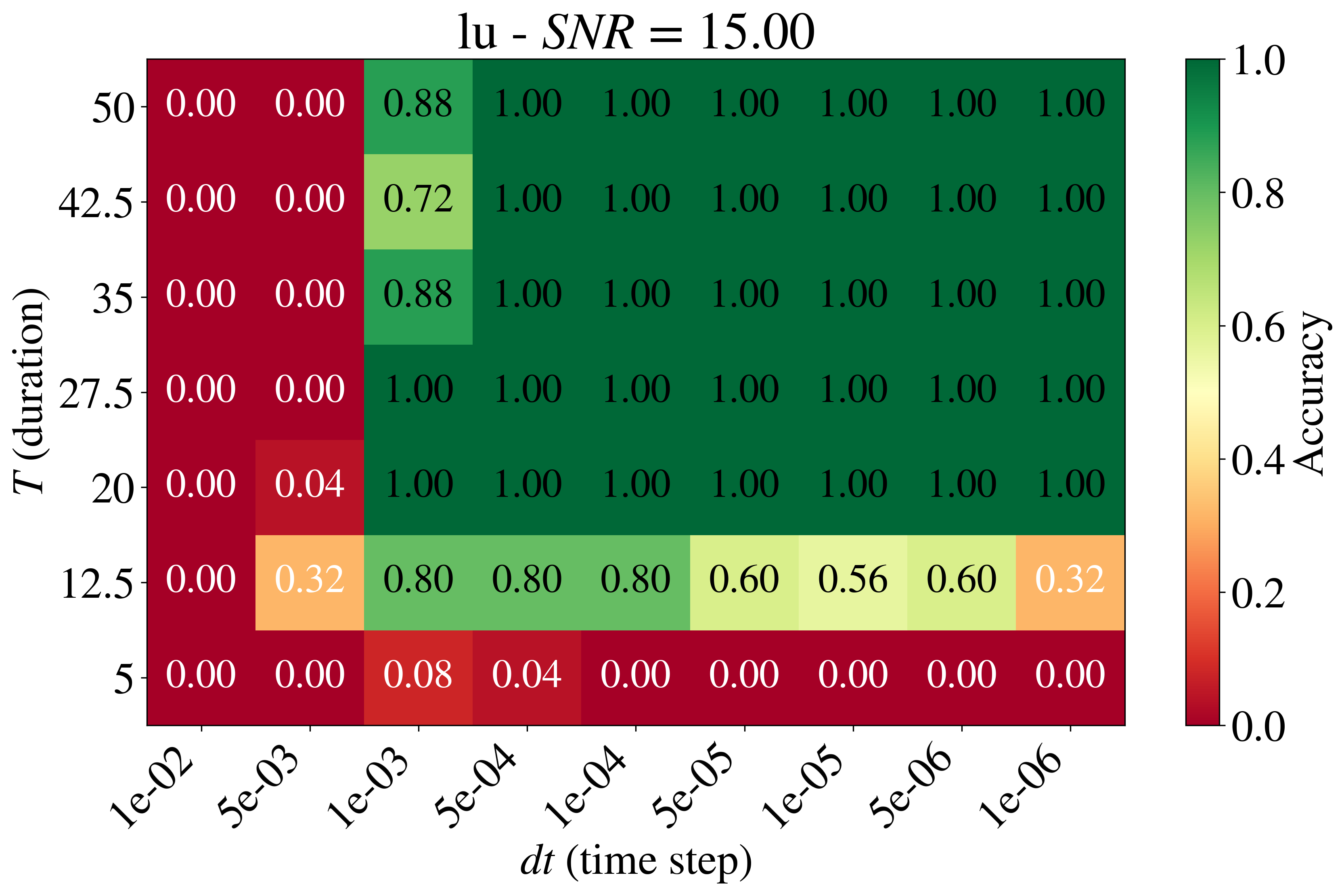}
\includegraphics[width=0.49\linewidth]{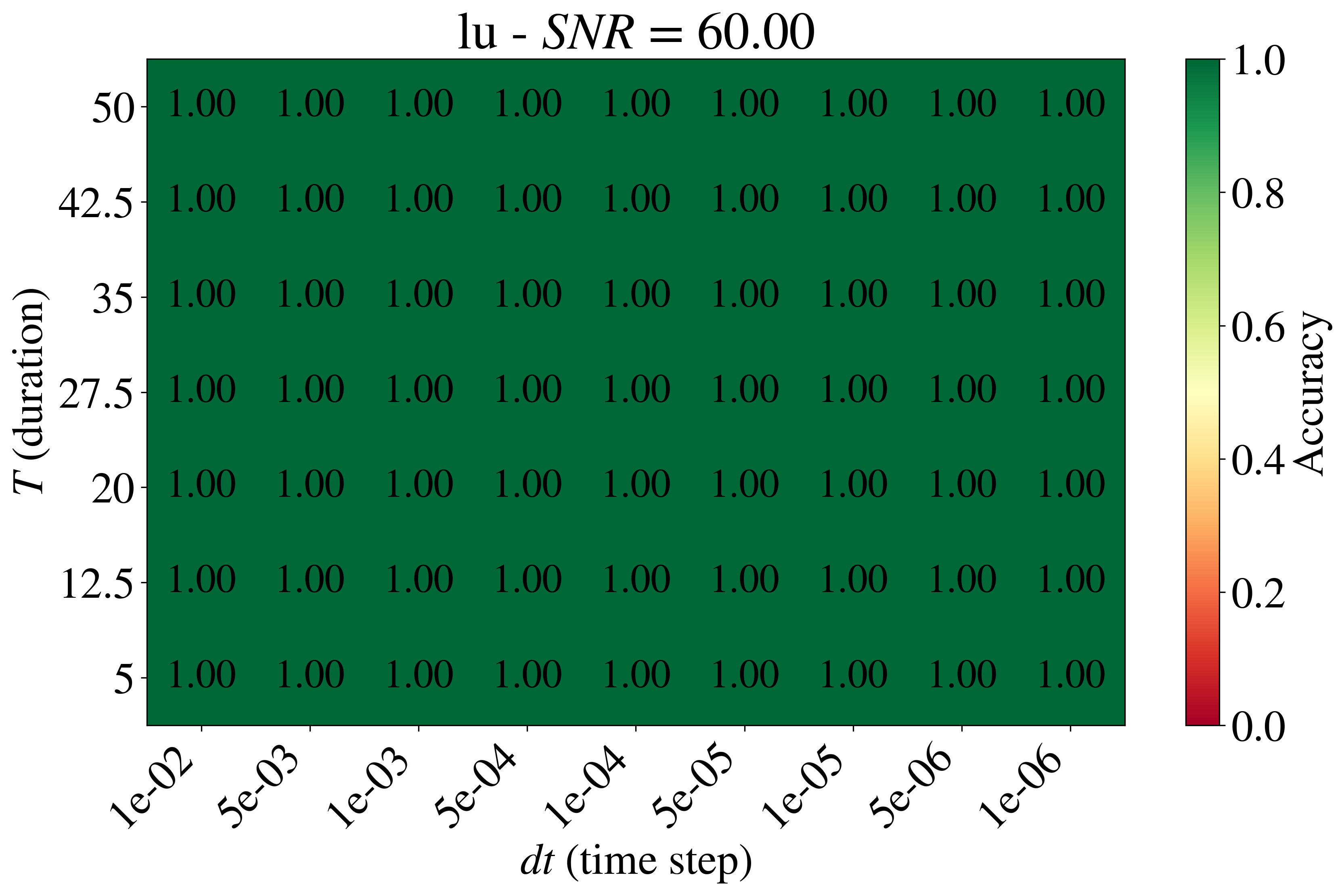}
\caption{Heatmaps of accuracy for Chen and Lu systems, across various $SNR$.}
\label{fig:hmap_chen_lu}
\end{figure}

\begin{figure}[!htbp]
\centering
\includegraphics[width=0.49\linewidth]{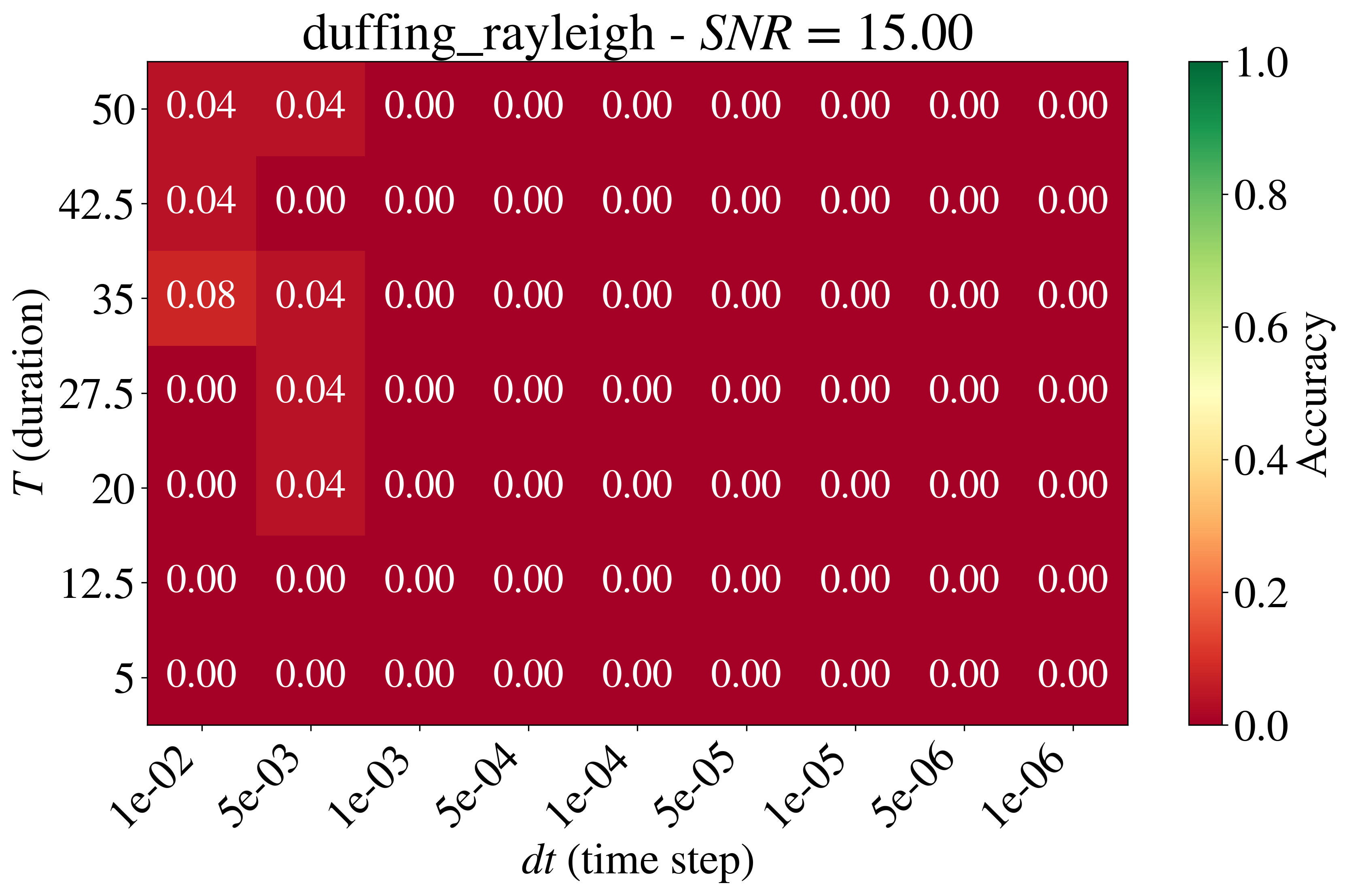}
\includegraphics[width=0.49\linewidth]{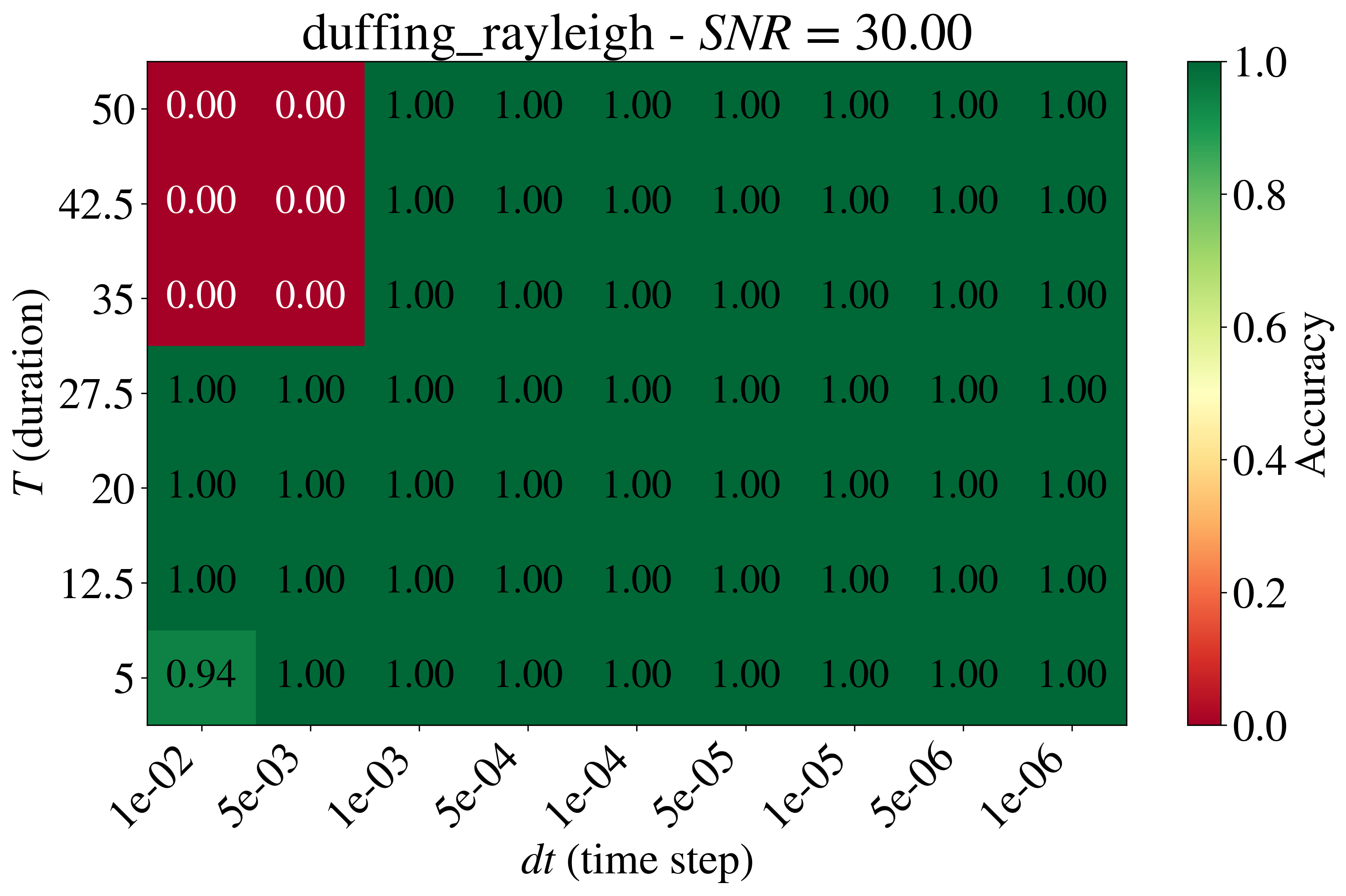}
\includegraphics[width=0.49\linewidth]{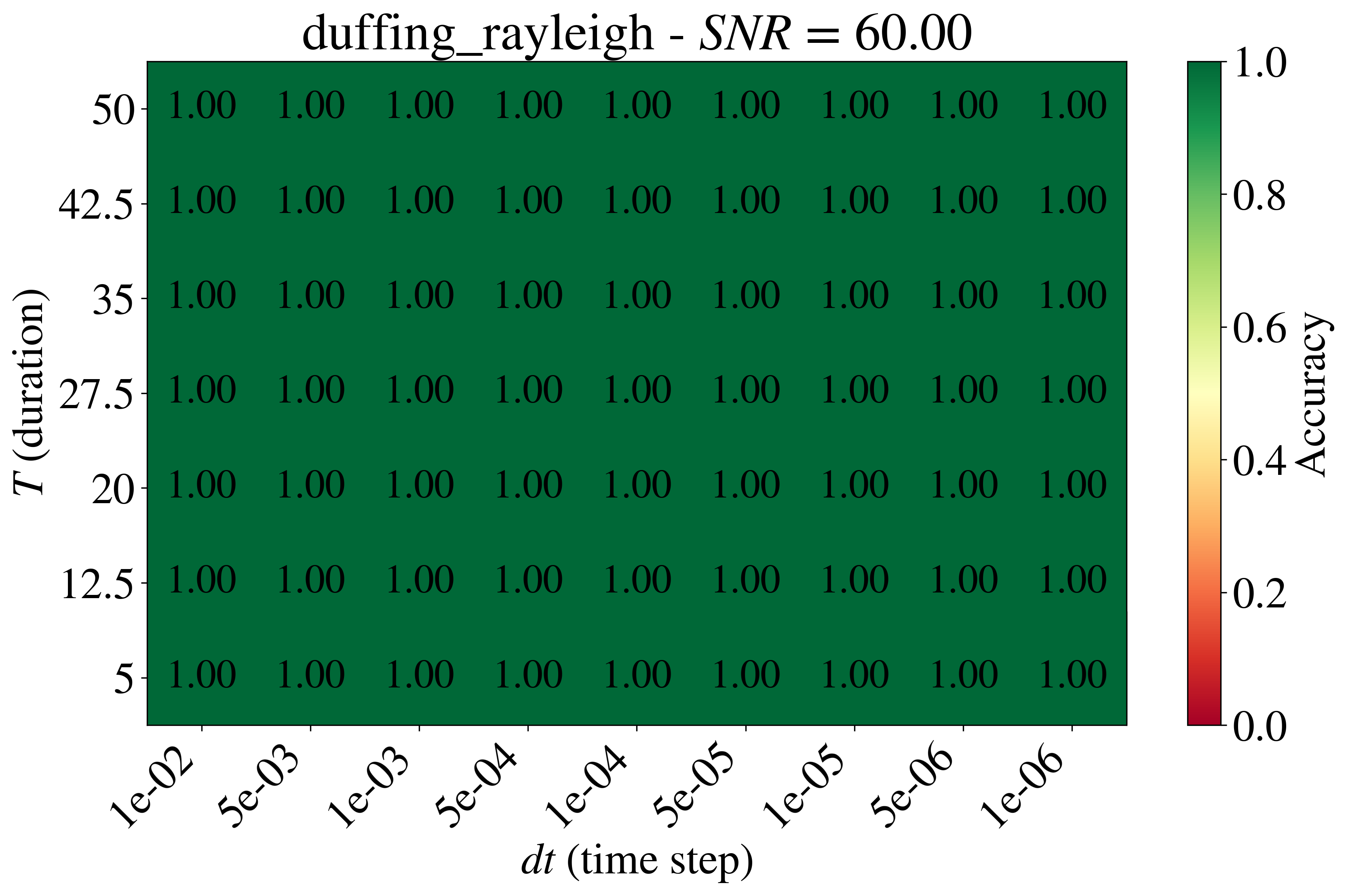}
\includegraphics[width=0.49\linewidth]{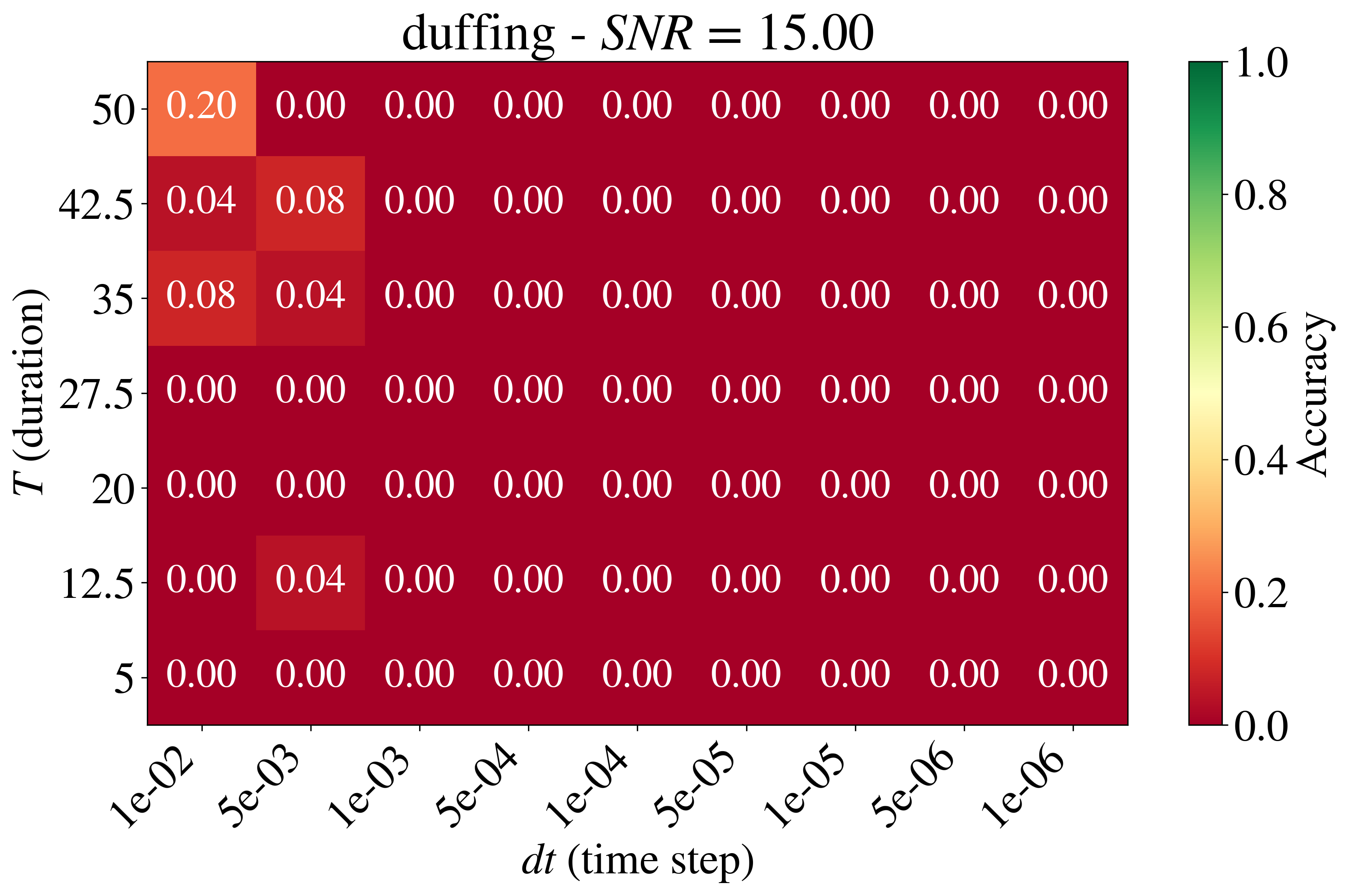}
\includegraphics[width=0.49\linewidth]{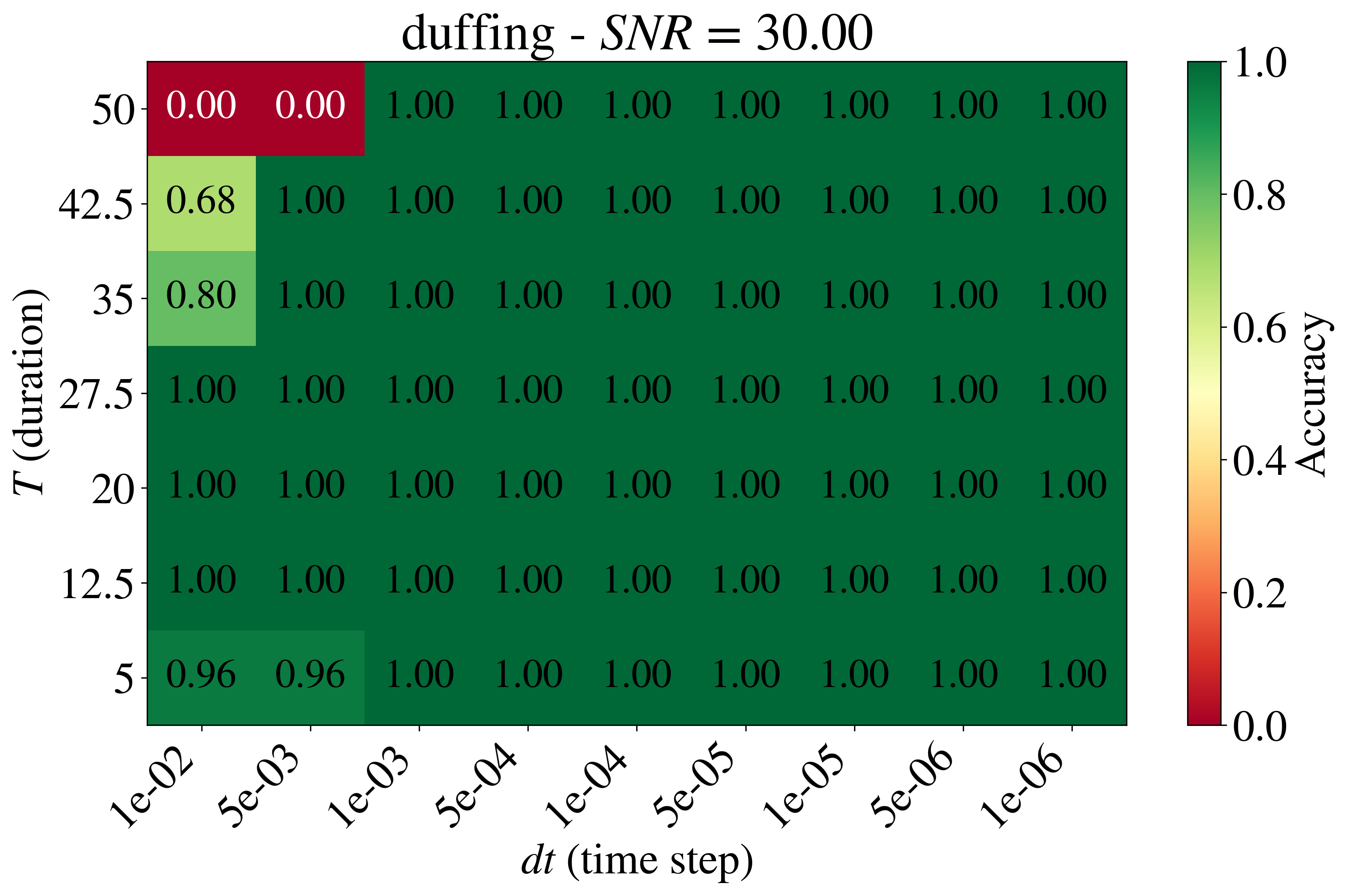}
\includegraphics[width=0.49\linewidth]{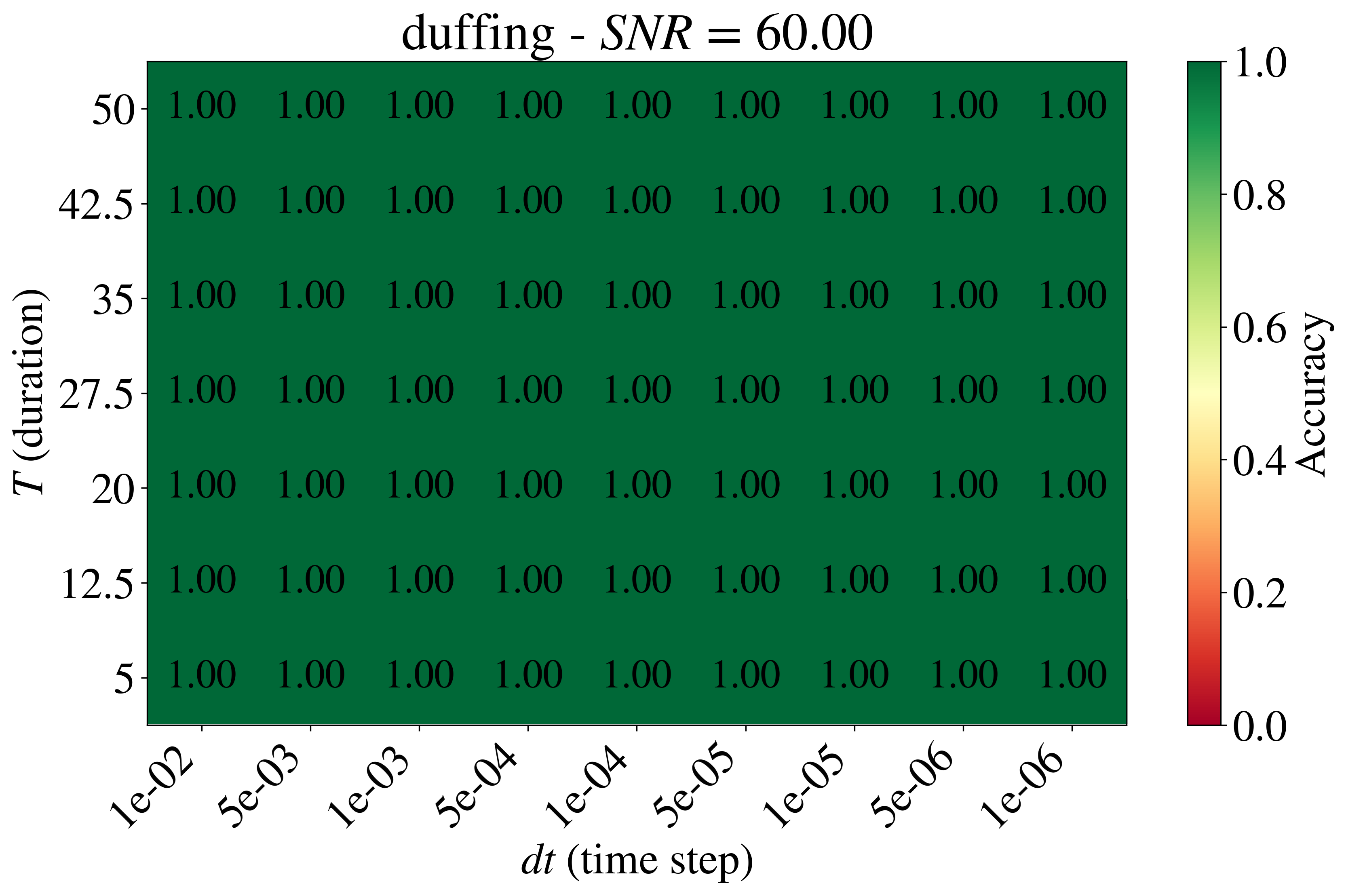}
\caption{Heatmaps of accuracy for Duffing and Rayleigh-Duffing systems, across various $SNR$.}
\label{fig:hmap_duffing}
\end{figure}

We next consider deterministic chaotic systems including duffing, rayleigh-duffing, chen and lu. Fig.~\ref{fig:hmap_chen_lu} shows heatmaps of accuracy for chen and lu systems. Both systems behave similarly, and their classification is reasonably accurate even for low SNRs---granted $dt$ and $T$ are sufficiently large. Fig.~\ref{fig:hmap_duffing} shows the accuracy heatmaps for duffing and rayleigh-duffing systems; with both systems observing the same trend---high classification accuracy beyond a critical SNR.

\subsection{Stochastic Chaotic Systems}

\begin{figure}[!htbp]
\centering
\includegraphics[width=0.65\linewidth]{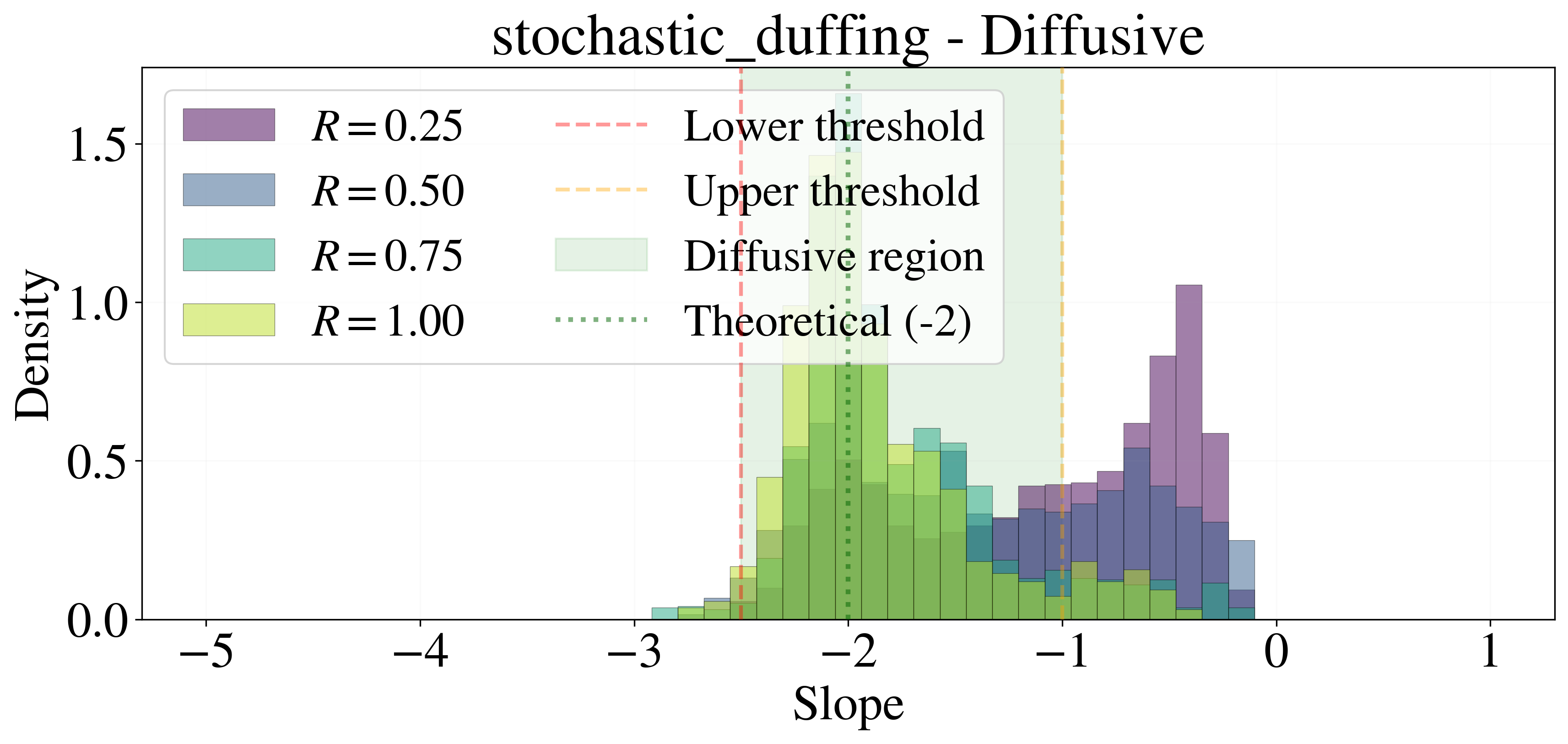}
\includegraphics[width=0.49\linewidth]{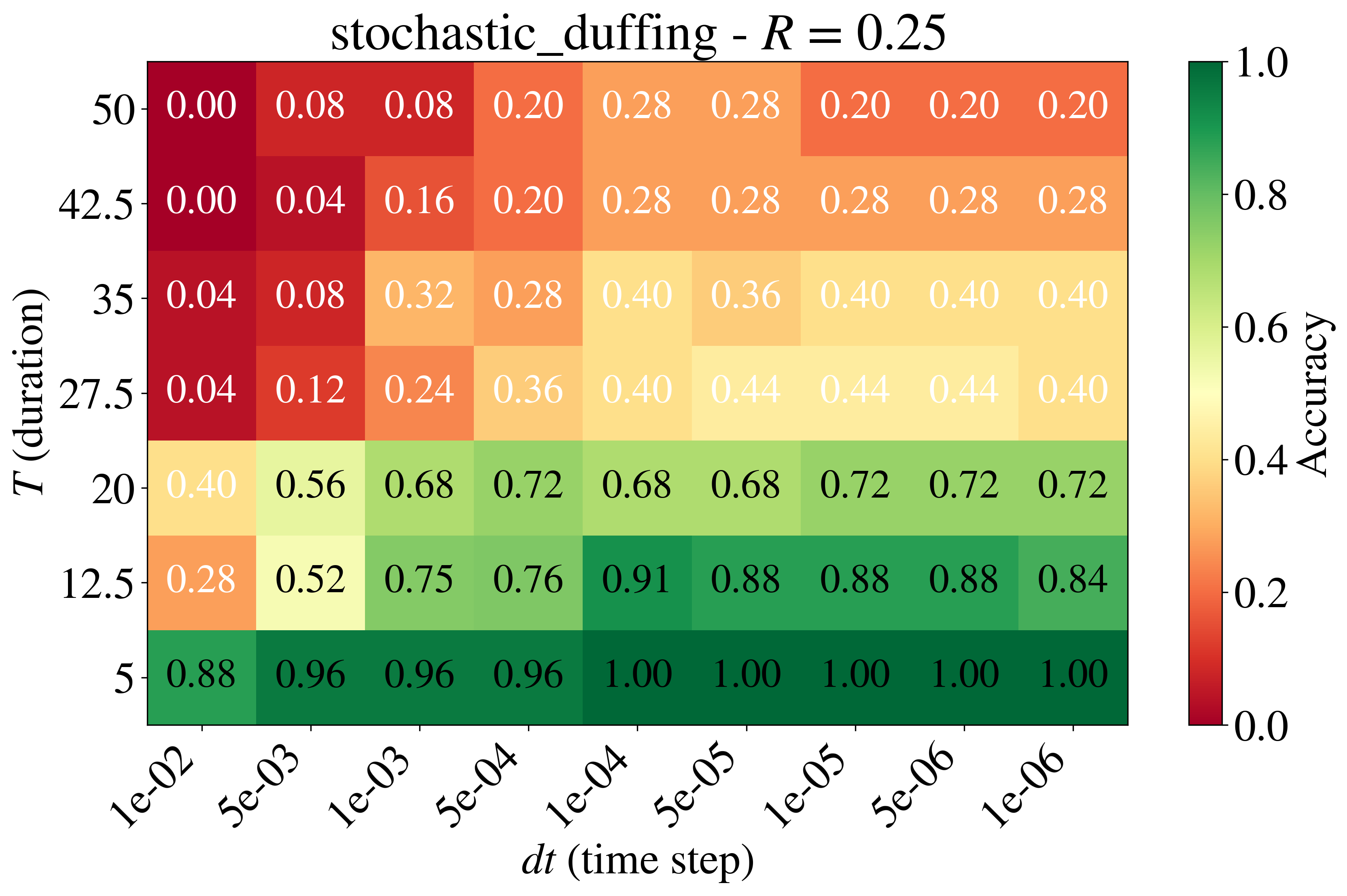}
\includegraphics[width=0.49\linewidth]{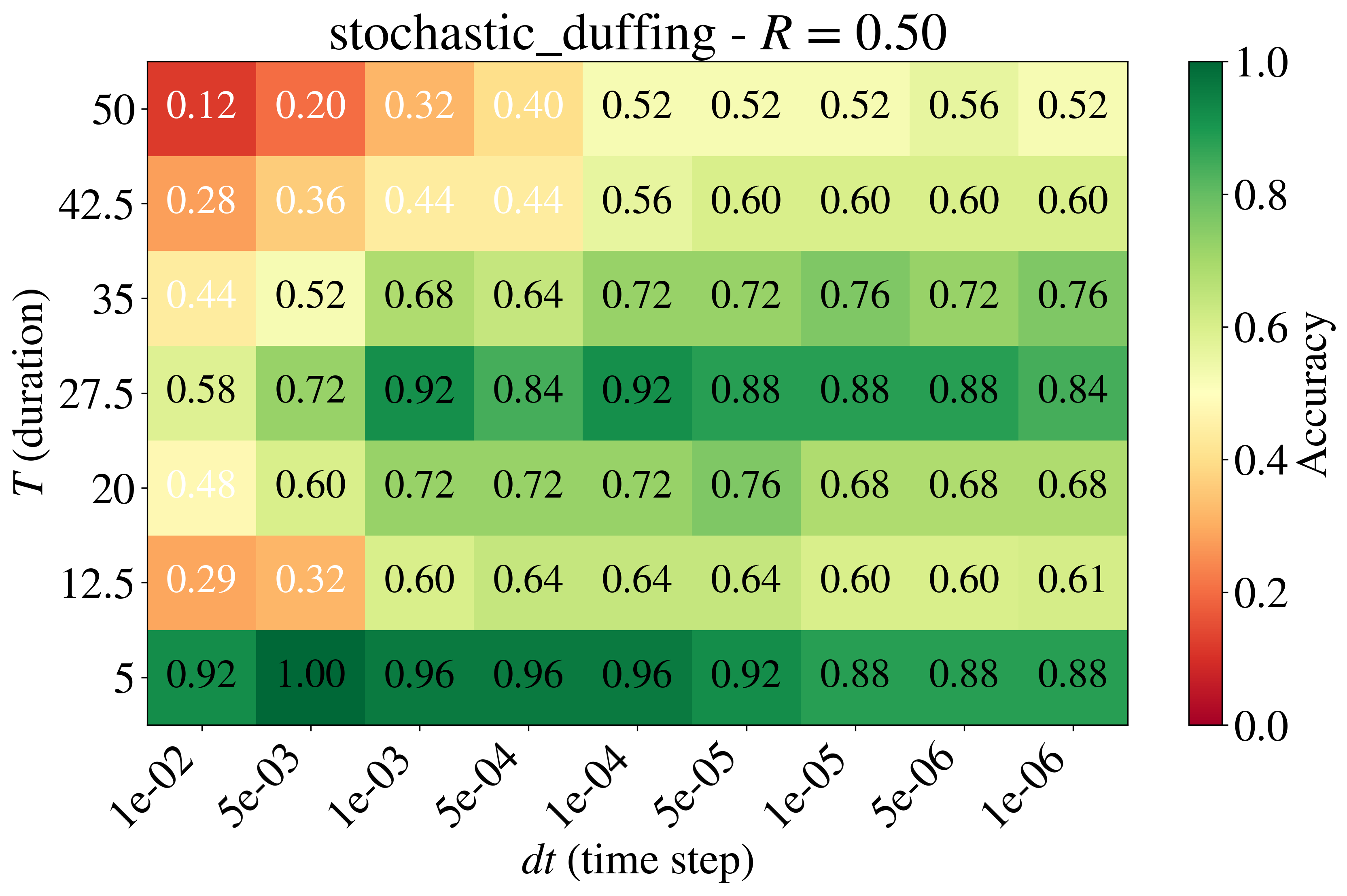}
\includegraphics[width=0.49\linewidth]{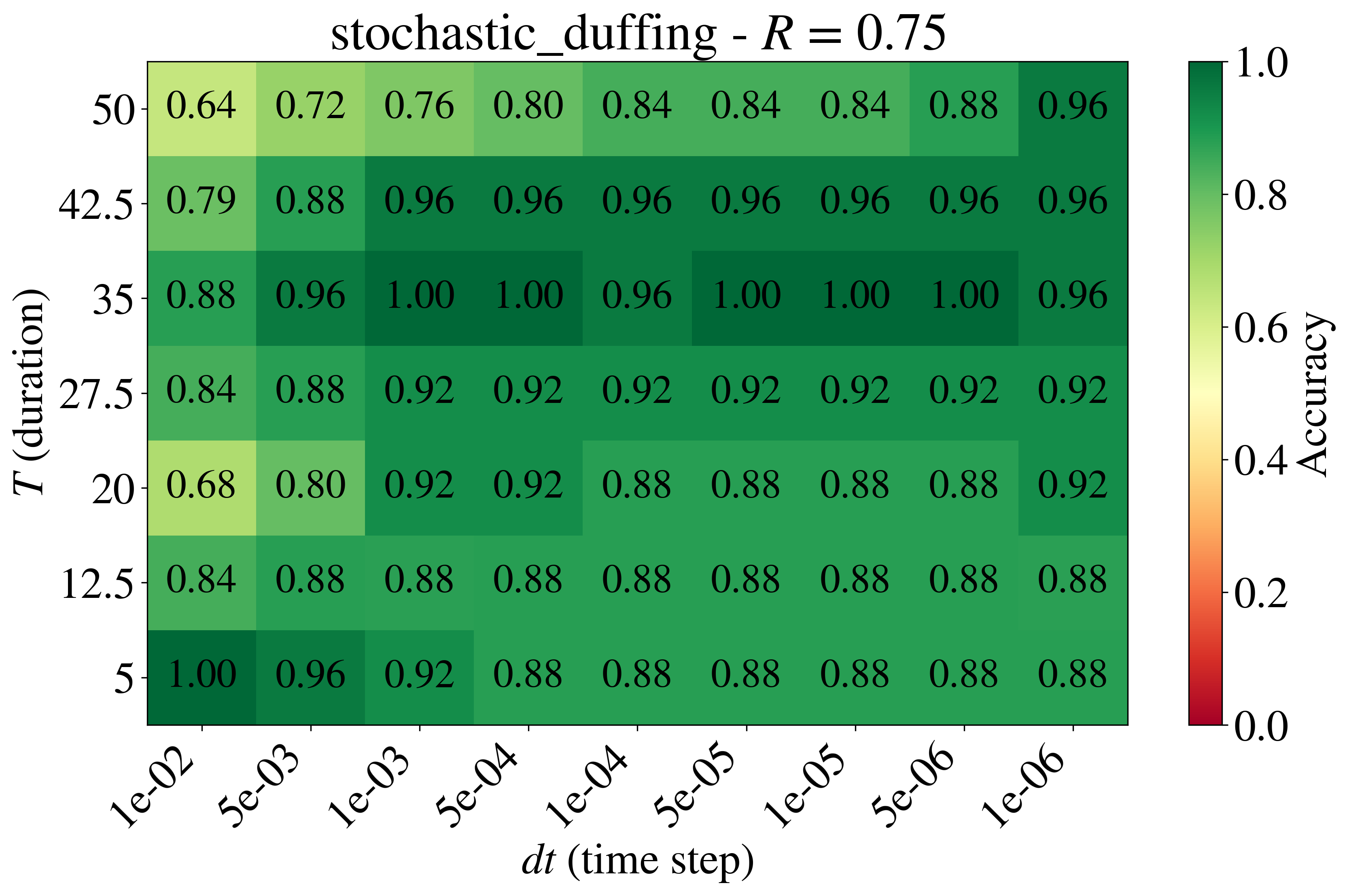}
\includegraphics[width=0.49\linewidth]{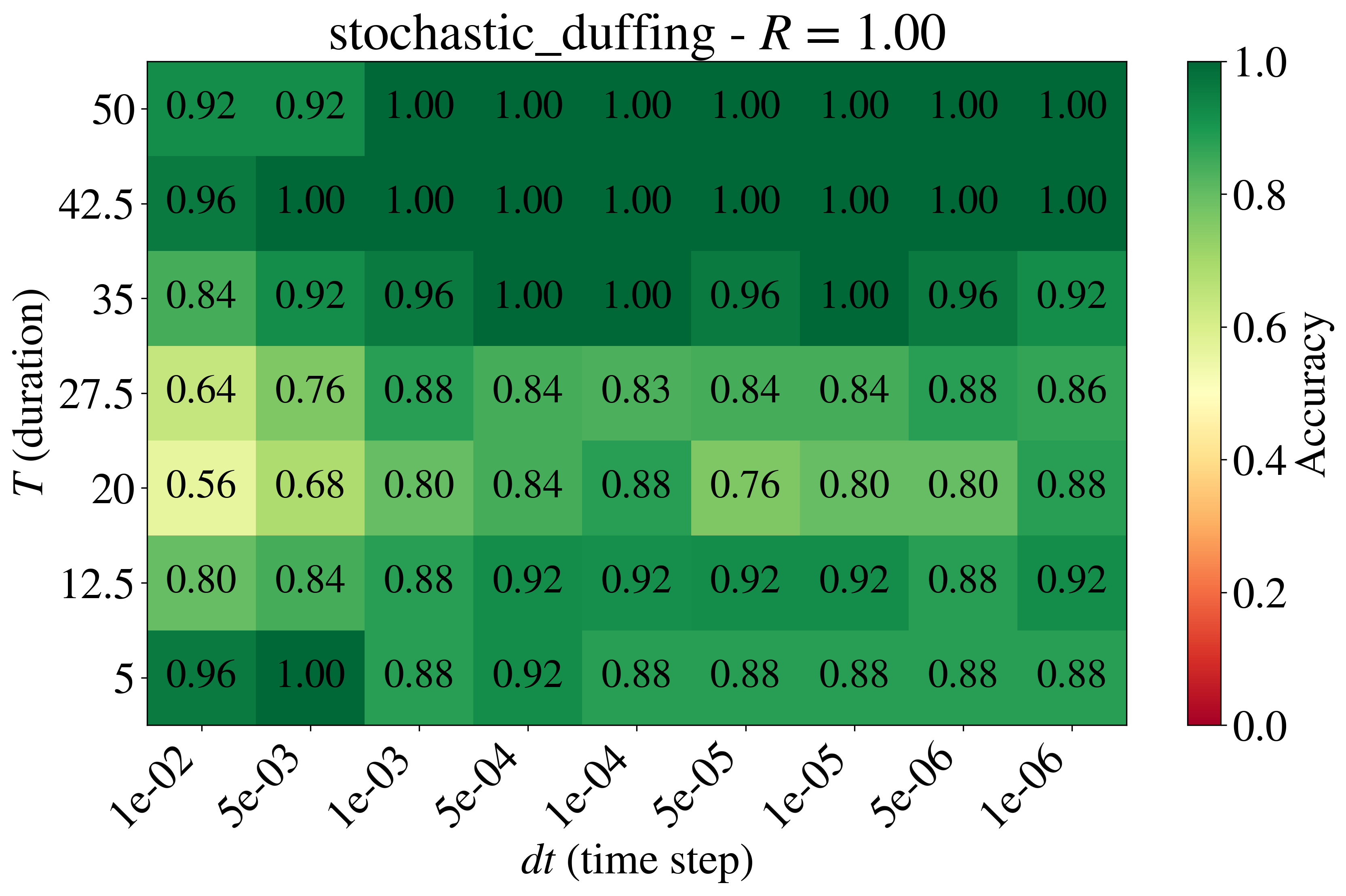}
\caption{Slope distribution and heatmaps of accuracy for stochastic Duffing system, across various $R$.}
\label{fig:stochastic_duffing}
\end{figure}

Further, we examine a deterministic chaotic system with intrinsic stochastic forcing---the stochastic Duffing oscillator. The stochastic Duffing system exhibits a clear monotonic increase in diffusion classification accuracy as noise intensity increases. Correspondingly, slope histograms narrow toward $-2$, indicating convergence toward diffusive behavior. See Fig.~\ref{fig:stochastic_duffing} for slope distributions and heatmaps of accuracy.

\subsection{Real-World Datasets}
\subsubsection{Financial Time Series}

\begin{figure}[!htbp]
    \centering
    \includegraphics[width=0.47\linewidth]{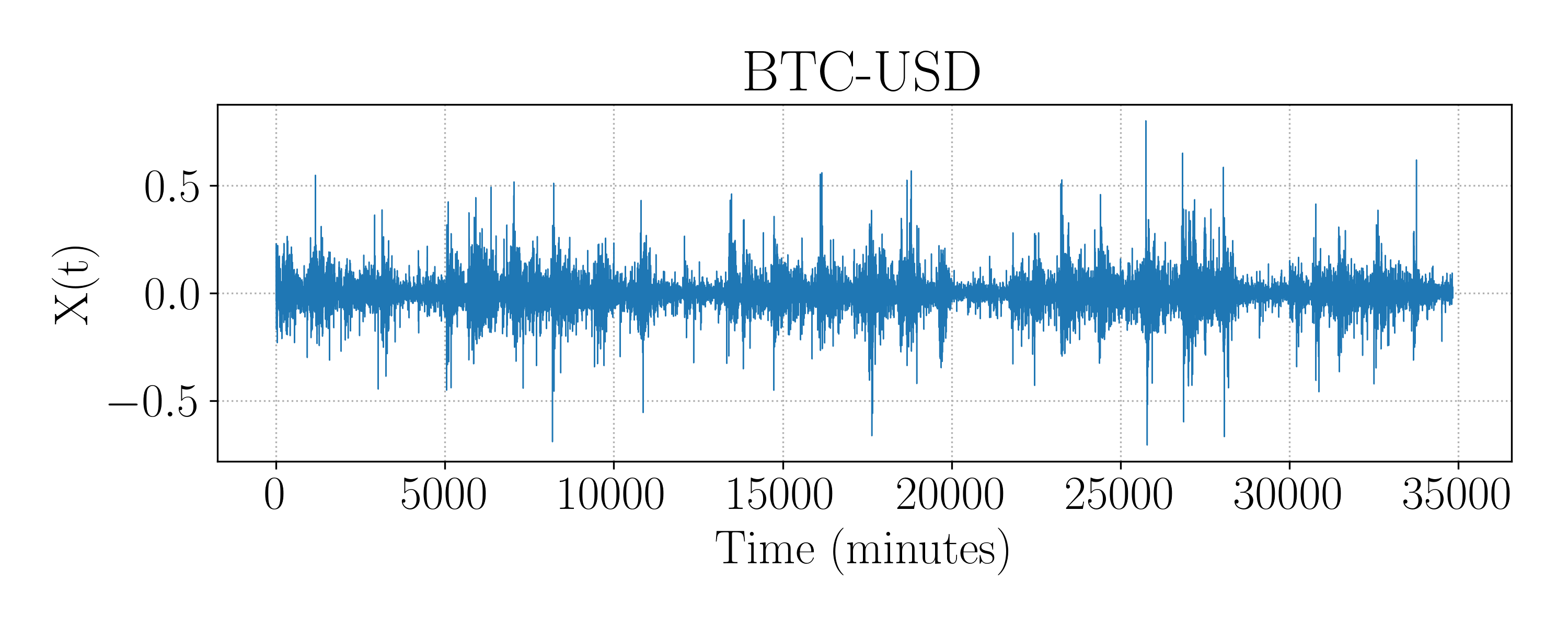}
    \includegraphics[width=0.47\linewidth]{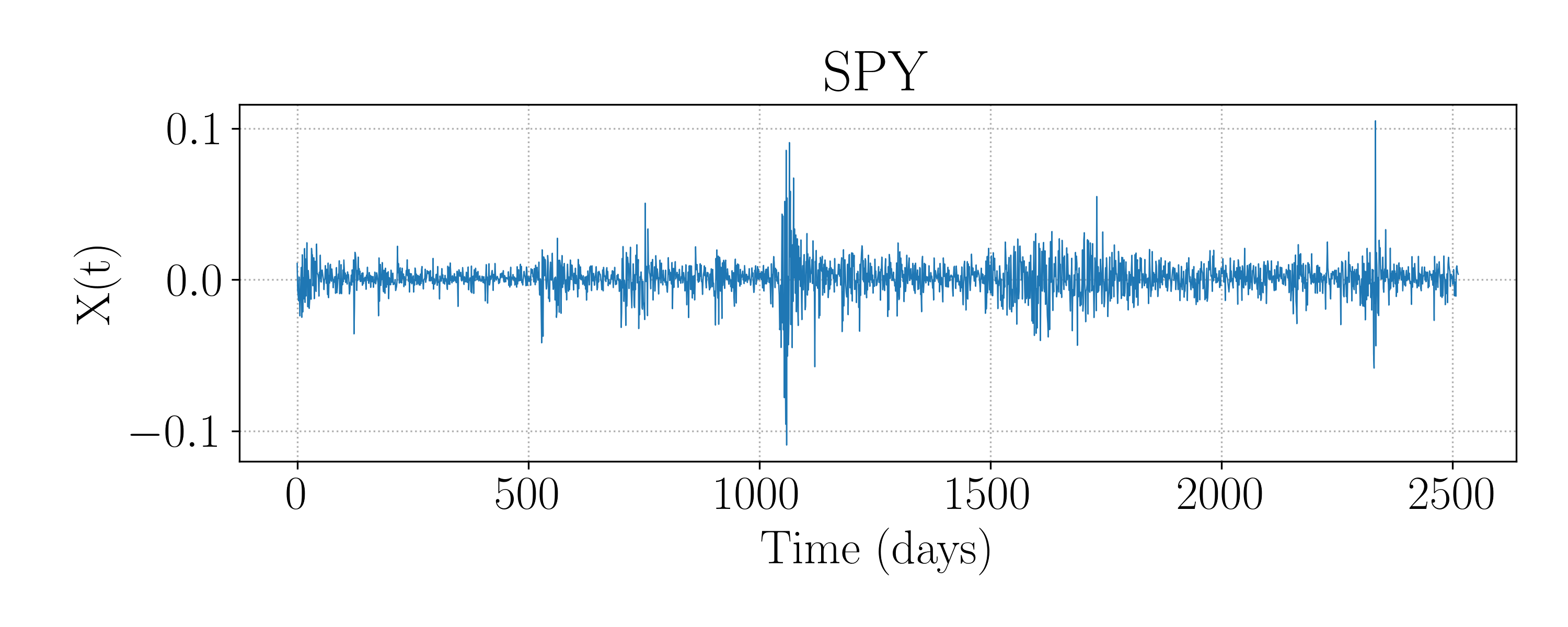}
    \caption{BTC-USD and SPY financial time series retrieved from yfinance on December 25, 2025.}
    \label{fig:finance}
\end{figure}

To demonstrate practical utility in real observational settings, we also applied the proposed method to financial market data, where differentiating diffusion-driven price variability from deterministic predictability is of fundamental interest. We analyzed minute-resolution returns of Bitcoin (BTC-USD) over a 30-day period, and daily returns of the S\&P 500 exchange-traded fund (SPY) over the past ten years, see signals in Fig.~\ref{fig:finance}. For each dataset, empirical returns were computed as
\[
r_t = \frac{P_{t} - P_{t-\Delta t}}{P_{t-\Delta t}},
\]
and treated as discrete observations of a stochastic process with sampling interval $\Delta t = 1$ minute (BTC) or $\Delta t = 1$ day (SPY).

For both assets, the fitted log--log slopes of the excursion count scaling law fell within the diffusion regime. Specifically, $\text{BTC: } \; \hat{s} \approx -2.11, \text{ and SPY: } \; \hat{s} \approx -1.61$. Both values fall within $-2.5 < \hat{s} < -1.0$, indicating the presence of a nontrivial local martingale component. The Bitcoin data shows behavior closely aligned with Brownian scaling, while SPY exhibits mild deviations possibly due to the lower discretization.

Importantly, our method achieves these results without any assumptions of stationarity, absence of drift, or Gaussian increments. Unlike model-dependent diffusion diagnostics, the excursion test observes path characteristics---enabling a direct conclusion: at the scales observed, both BTC-USD and SPY returns possess locally diffusive structure with finite quadratic variation.

\subsubsection{Real-world Audio Signals}

We further evaluated the proposed excursion-based diffusion detector on a diverse real-world dataset of short audio signals~\cite{ESC}. The dataset contained 50 commonly studied sound classes (environmental, human, mechanical, and animal), each represented by 40 independent samples of 5 seconds duration each, sampled at uniform audio rate of 44.1 kHz. Amplitude envelopes of each were treated as the observed trajectory $X_t$, and each of the independent samples was classified as either diffusive or non-diffusive using our algorithm. Since there is no known ground truth for any of these audio signals, we compute and report the percentage of samples which got classified as diffusive for each sound type. 

Diffusive behaviour was detected in a majority of sounds, including natural ambience such as {rain} (for which 82.5\% of samples were classified diffusive), {thunderstorm} (90.0\%), {wind} (77.5\%), {sea waves} (97.5\%), and biological vocalizations such as {sheep bleats} (92.5\%), {pig grunts} (95.0\%), and {frogs} (80.0\%). Anthropogenic noise sources also frequently get classified: {vacuum cleaner} (95.0\%), {engine running} (92.5\%), {chainsaw} (100.0\%), {train} (95.0\%), and {helicopter} (87.5\%). These results align with the presence of broadband stochastic components in aerodynamic and combustive processes.

More structured or transient-dominated signals show reduced diffusion classifications: {mouse clicks} (52.5\%), {clock tick} (50.0\%), {can opening} (50.0\%), {glass breaking} (70.0\%), and {church bells} (97.5\%) occupy a broad continuum between deterministic periodicity and random excitation. Vocal sounds such as {laughing} (82.5\%), {crying baby} (95.0\%), and {coughing} (90.0\%) are also predominantly classified diffusive, reflecting turbulent airflow in speech production.

Interestingly, simple and rhythmic event sequences like {clock alarm} (45.0\%) and {door knock} (37.5\%) exhibit less diffusive signatures likely due to repetitive deterministic structure. Overall, the results demonstrate that the proposed method is capable of distinguishing fine-grained differences in physical processes. Even across complex, real-world acoustic environments, signals exhibiting turbulent or noisy dynamics systematically trigger the characteristic $N(\varepsilon) \sim \varepsilon^{-2}$ scaling behavior, while impulsive or organized sounds do not. Therefore, this test provides a model-free pathway to partition high-dimensional multimedia signals into diffusion-driven versus deterministic regimes, offering potential applications in many fields.

\section{Limitations}

The proposed excursion-scaling test applies broadly to continuous semimartingales with finite quadratic variation. However, several practical considerations limit its universal deployment. First, reliable excursion statistics require fine resolution: large sampling intervals $dt$ suppress small excursions and bias the estimate toward deterministic behavior. Similarly, low signal-to-noise environments can cause a truly deterministic signal to be misclassified. The method does not apply to jump processes, impulsive dynamics, or Levy-type models with infinite-activity discontinuities, since their excursion structure falls outside the theoretical regime. Fractional Brownian motion also lacks the finite quadratic variation needed for correct scaling. These constraints highlight that the method performs best when the underlying dynamics are continuous in time, well-resolved by sampling, and maintain sufficient variance to reveal the scaling law predicted by semimartingale theory.

\section{Conclusion}

We introduced a principled excursion-asymptote law for continuous semimartingales of Ito type, and demonstrated its use as a robust diagnostic to distinguish stochastic diffusion from deterministic/chaotic dynamics. Unlike many existing chaos-versus-noise discrimination methods that require extensive hyperparameter tuning, multiple trained models, or subjective interpretations of surrogate metrics, our approach is fully self-contained and leverages a theoretically guaranteed scaling law for excursion counts. Across a broad benchmark of systems, the method exhibits strong performance---achieving near-perfect accuracy on canonical stochastic processes such as Brownian motion, Ornstein-Uhlenbeck dynamics, and Cox-Ingersoll-Ross diffusion. Deterministic oscillators and discrete maps are reliably classified when the time resolution is sufficiently fine, the time horizon is large enough to capture attractor dynamics, and the signal-to-noise ratio remains moderate to high. Stochastic chaotic systems are also correctly identified over a wide range of noise intensities when temporal resolution and duration are adequate. On real-world financial data, including minute-level cryptocurrency returns and daily equity returns, the method perfectly identified diffusive behavior consistent with market theory. Results on short environmental and human-generated audio clips further reveal interesting distinctions in how natural signals manifest stochasticity. Overall, these findings highlight the strong practical potential of the proposed diagnostic as a simple, theory-driven tool for separating stochasticity from chaotic determinism in physical, biological, and engineered systems.

\section{Acknowledgements}
This material is based upon work supported by the Air Force Office of Scientific Research under award number FA9550-26-1-0011, and by the NSF Frontera Computational Science Fellowship awarded to ST for 2025-2026. 

\printbibliography
\appendix
\section{Benchmark System Definitions}
\label{appendix:systems}

All benchmark signals used in this study are generated as one-dimensional
observables from standard deterministic chaotic systems, continuous-time
stochastic differential equations (SDEs), and hybrid stochastic-chaotic
dynamics. Deterministic ODEs are integrated using adaptive Bogacki-Shampine
RK23 with fixed output grid~$dt$, while all SDEs are discretized using
Euler-Maruyama. Maps are iterated directly. In every case, only the first
state component $x(t)$ is recorded as the output. Here, $W_t$ denotes a standard Wiener process, and $R>0$ scales noise intensity via $\sigma = 1/R$.

\paragraph{Simple harmonic oscillator:}
\begin{equation}
\ddot{x} + \omega^2 x = 0,
\qquad \omega = 1.0.
\end{equation}

\subsection{Chaotic Attractors}

\paragraph{Duffing oscillator:}
\begin{align}
\dot{x} &= v, \\
\dot{v} &= -\delta v - \alpha x - \beta x^3 + \gamma \cos(\omega t),
\end{align}
with $\delta = 0.2,\ \alpha=-1,\ \beta = 1,\ \gamma = 0.3,\ \omega = 1.2$.

\paragraph{Rayleigh-Duffing oscillator:}
\begin{align}
\dot{x} &= v, \\
\dot{v} &= -\delta v - \varepsilon \left(v - v^3\right)
         - \alpha x - \beta x^3 
         + \gamma \cos(\omega t),
\end{align}
with parameters $\delta = 0.2, \alpha = -1, \beta = 1, \gamma = 0.3, \omega = 1.2, \varepsilon = 0.1$.

\paragraph{Chen attractor:}
\begin{align}
\dot{x} &= a(y-x), \\
\dot{y} &= (c-a)x - xz + cy, \\
\dot{z} &= xy - bz,
\end{align}
with $(a,b,c) = (35,\, 3,\, 28)$.

\paragraph{Lu attractor:}
\begin{align}
\dot{x} &= a(y-x), \\
\dot{y} &= -xz + cy, \\
\dot{z} &= xy - bz,
\end{align}
with $(a,b,c) = (36,\, 3,\, 20)$.

\subsection{Chaotic Maps}

For all maps, $x_{k+1}$ is observed at discrete times $t_{k+1}=t_k+dt$.

\paragraph{Logistic map:}
\begin{equation}
x_{k+1} = r x_k (1 - x_k), \qquad r = 4.0.
\end{equation}

\paragraph{Henon map:}
\begin{align}
x_{k+1} &= 1 - a x_k^2 + y_k, \\
y_{k+1} &= b x_k,
\end{align}
with $(a,b) = (1.4,\, 0.3)$.

\paragraph{Linear Congruential Generator (LCG):}
\begin{equation}
x_{k+1} = (a x_k + c) \bmod m,
\qquad
a = 1664525,\ 
c = 1013904223,\ 
m = 2^{32}.
\end{equation}

\subsection{Stochastic Diffusions}

These are simulated as Ito SDEs,
\[
dX_t = f(X_t,t)\,dt + g(X_t,t)\,dW_t,\qquad g=\sigma=\frac{1}{R}.
\]

\paragraph{Brownian motion:}
\begin{equation}
dX_t = \sigma \, dW_t.
\end{equation}

\paragraph{Ornstein-Uhlenbeck (OU) process:}
\begin{equation}
dX_t = \theta (\mu - X_t)\,dt + \sigma\, dW_t,
\qquad \theta = 1,\ \mu = 0.
\end{equation}

\paragraph{Cox-Ingersoll-Ross (CIR) process:}
\begin{equation}
dX_t = \kappa(\theta - X_t)\,dt + \sigma \sqrt{X_t}\,dW_t,
\qquad X_t \ge 0.
\end{equation}

\paragraph{Stochastic Duffing:}
\begin{align}
dX &= V\,dt + \sigma\, dW_t, \\
dV &= \bigl(-0.2V + X + X^3 + 0.3\cos(1.2t)\bigr)dt + \sigma\, dW_t,
\end{align}
using the same scalar Wiener increment in both equations (rank-1 diffusion).

\end{document}